\RequirePackage{fix-cm}
\documentclass[twocolumn,natbib]{svjour3}          %
\smartqed  %
\usepackage{graphicx}
\usepackage{mathptmx}      %
\clearpage{}%
\makeatletter{}%
\usepackage{color,xcolor}
\usepackage{epsfig}
\usepackage{graphicx}

\usepackage{adjustbox}
\usepackage{array}
\usepackage{booktabs}
\usepackage{colortbl}
\usepackage{float,wrapfig}
\usepackage{hhline}
\usepackage{multirow}
\usepackage{bm}
\usepackage{nicefrac}
\usepackage{changepage}
\usepackage{extramarks}
\usepackage{fancyhdr}
\usepackage{lastpage}
\usepackage{setspace}
\usepackage{soul}
\usepackage{xspace}
\usepackage{subfig}

\usepackage{url}

\usepackage{algorithm, algorithmic}
\usepackage{enumerate}

\usepackage{comment}
\usepackage{mathtools}
\usepackage[pagebackref=true,breaklinks=true,colorlinks,citecolor=blue,bookmarks=false]{hyperref}
\clearpage{}%
\clearpage{}%
\makeatletter{}%

\newcommand{\sect}[1]{Section~\ref{#1}}

\newcommand{\vpar}[1]{\paragraph{\normalfont\bf #1}\ \ }
\newcommand{\xpar}[1]{\paragraph{\normalfont\bf #1}\ \ }

\newcommand{\fig}[1]{Figure~\ref{#1}}
\newcommand{\tbl}[1]{Table~\ref{#1}}
\newcommand{\ignorethis}[1]{}

\newcommand{\eg}{{e.g.}\@\xspace}
\newcommand{\ie}{{i.e.}\@\xspace}

\newcommand{\binmod}{Binary\xspace}
\newcommand{\clustermod}{Clustering\xspace}
\newcommand{\spectmodel}{Spectrum\xspace}

\usepackage{array}
\newcolumntype{L}[1]{>{\raggedright\let\newline\\\arraybackslash\hspace{0pt}}m{#1}}
\newcolumntype{C}[1]{>{\centering\let\newline\\\arraybackslash\hspace{0pt}}m{#1}}
\newcolumntype{R}[1]{>{\raggedleft\let\newline\\\arraybackslash\hspace{0pt}}m{#1}}

\clearpage{}%
\clearpage{}%
\makeatletter{}%
\newcommand{\winsizesec}{3.75\xspace}
\newcommand{\numfullbands}{32\xspace}

\newcommand{\numpervideo}{10\xspace}

\newcommand{\soundtexdim}{502}

\newcommand{\numclusters}{30\xspace}
\newcommand{\testclusteracc}{15.8\%\xspace}
\newcommand{\clusterpurechance}{3.3\%\xspace}
\newcommand{\clustermostcommon}{6.6\%\xspace}
\newcommand{\numtrainimsmil}{1.8\xspace} %

\newcommand{\numobjsununits}{67\xspace}
\newcommand{\numobjtrackingsununits}{61\xspace}

\newcommand{\numobjunits}{91\xspace}
\newcommand{\numaudiosetunits}{84\xspace}
\newcommand{\numplacesunits}{117\xspace}
\newcommand{\numtrackingunits}{72\xspace}
\newcommand{\nummotionunits}{27\xspace}
\newcommand{\fracvidsound}{43.7\%\xspace}

\clearpage{}%
\journalname{Int J Comput Vis}
\begin{document}

\title{Learning Sight from Sound: \\Ambient Sound Provides Supervision for Visual Learning}

\author{Andrew Owens \and
Jiajun Wu \and
Josh H. McDermott \and
William T. Freeman \and\\
Antonio Torralba 
}

\institute{Andrew Owens \at
  University of California, Berkeley \\
  Massachusetts Institute of Technology \\
              \email{andrewo@mit.edu}           %
           \and
           Jiajun Wu \at
              Massachusetts Institute of Technology \\
              \email{jiajunwu@mit.edu}
           \and
           Josh H. McDermott \at
              Massachusetts Institute of Technology \\
              \email{jhm@mit.edu}
           \and
           William T. Freeman \at
           Massachusetts Institute of Technology \\
           Google Research \\
              \email{billf@mit.edu}
           \and
           Antonio Torralba \at
              Massachusetts Institute of Technology \\
              \email{torralba@csail.mit.edu}
}

\date{Received: date / Accepted: date}

\maketitle

\makeatletter{}%
\begin{abstract}
The sound of crashing waves, the roar of fast-moving cars~--~sound conveys important information about the objects in our surroundings. In this work, we show that ambient sounds can be used as a supervisory signal for learning visual models.  To demonstrate this, we train a convolutional neural network to predict a statistical summary of the sound associated with a video frame.  We show that, through this process, the network learns a representation that conveys information about objects and scenes.  We evaluate this representation on several recognition tasks, finding that its performance is comparable to that of other state-of-the-art unsupervised learning methods. Finally, we show through visualizations that the network learns units that are selective to objects that are often associated with characteristic sounds. This paper extends an earlier conference paper, \citet{owens2016ambient}, with additional experiments and discussion.
\end{abstract}
\makeatletter{}%
\section{Introduction}

\makeatletter{}%

\begin{figure*}
  {\centering
  \includegraphics[width=1.00\linewidth]{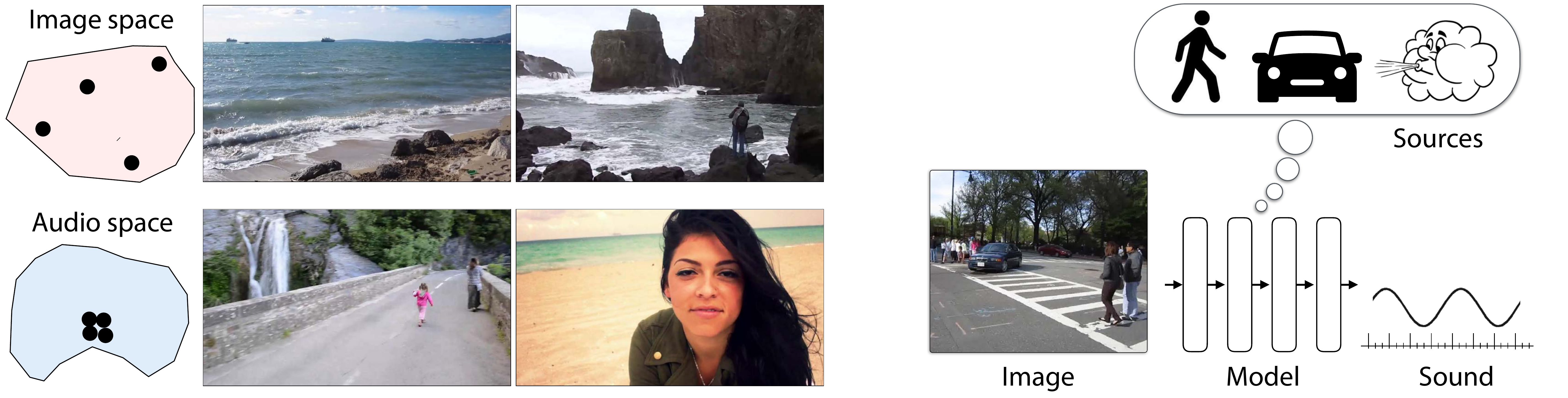}\vspace{2mm}}
  \indent \hspace{20mm} (a) Videos with similar audio tracks
  \hspace{45mm} (b) Audio-visual structures
  \caption{\label{fig:motivation} Predicting audio from images
    requires an algorithm to generalize over a variety of visual
    transformations. During the learning process, a sound-prediction
    algorithm will be forced to explain why the images in (a) are
    closely clustered in audio feature space.  This requires detecting
    the water in the scene (or its close correlates, such as sand),
    while ignoring variations in appearance -- the scene illumination,
    the angle of the camera, the presence of people in frame -- that
    do not affect the sound.  A sound-prediction model thus must (b)
    learn to recognize structures that appear in both modalities.}
\end{figure*}

Sound conveys important information about the world around us -- the bustle of a caf\'{e} tells us that there are many people nearby, while the low-pitched roar of engine noise tells us to watch for fast-moving cars \citep{gaver1993world}. Although sound is in some cases complementary to visual information, such as when we listen to something out of view, vision and hearing are often informative about the same structures in the world. Here we propose that as a consequence of these correlations, concurrent visual and sound information provide a rich training signal that can be used to learn useful representations of the visual world.

In particular, an algorithm trained to predict the sounds that occur within a visual scene might be expected to learn about objects and scene elements that are associated with salient and distinctive noises, such as people, cars, and flowing water (\fig{fig:motivation}). Such an algorithm might also learn to associate visual scenes with the ambient sound textures \citep{mcdermott2011sound} that occur within them.  It might, for example, associate the sound of wind with outdoor scenes, and the buzz of refrigerators with indoor scenes.

Although human annotations are indisputably useful for learning, they are expensive to collect. The correspondence between ambient sounds and video is, by contrast, ubiquitous and free.  While there has been much work on learning from unlabeled image data \citep{doersch2015unsupervised,wang2015unsupervised,le2012building}, an audio signal may provide information that is largely orthogonal to that available in images alone -- information about semantics, events, and mechanics are all readily available from sound \citep{gaver1993world}.

One challenge in utilizing audio-visual input is that the sounds we hear are only loosely associated with what we see. Sound-producing objects often lie outside of our visual field, and objects that are capable of producing characteristic sounds -- barking dogs, ringing phones -- do not always do so. A priori it is thus not obvious what might be achieved by predicting sound from images.

In this work, we show that a model trained to predict held-out sound from video frames learns a visual representation that conveys semantically meaningful information.  We formulate our sound-prediction task as a classification problem, in which we train a convolutional neural network (CNN) to predict a statistical summary of the sound that occurred at the time a video frame was recorded.  We then validate that the learned representation contains significant information about objects and scenes.

We do this in two ways: first, we show that the image features that we learn through our sound-prediction task can be used for object and scene recognition. On these tasks, our features obtain performance that is competitive with that of state-of-the-art unsupervised and self-supervised learning methods. Second, we show that the intermediate layers of our CNN are highly selective for objects.  This augments recent work \citep{zhou2014object} showing that object detectors ``emerge'' in a CNN's internal representation when it is trained to recognize scenes.  As in the scene recognition task, object detectors emerge inside of our sound-prediction network. However, our model learns these detectors from an unlabeled audio-visual signal, without any explicit human annotation.

In this paper, we: (1) present a model based on visual CNNs and sound textures \citep{mcdermott2011sound} that predicts a video frame's held-out sound; (2) demonstrate that the CNN learns units in its convolutional layers that are selective for objects, extending the methodology of \cite{zhou2014object}; (3) validate the effectiveness of sound-based supervision by using the learned representation for object- and scene-recognition tasks.  These results suggest that sound data, which is available in abundance from consumer videos, provides a useful training signal for visual learning.

\makeatletter{}%
\section{Related Work}

We take inspiration from work in psychology, such as Gaver's Everyday
Listening \citep{gaver1993world}, that studies the ways that humans
learn about objects and events using sound.  In this spirit, we would
like to study the situations where sound tells us about visual objects
and scenes.  Work in auditory scene analysis
\citep{ellis2011classifying,eronen2006audio,lee2010detecting}
meanwhile has provided computational methods for recognizing
structures in audio streams.  Following this work, we use a sound
representation \citep{mcdermott2011sound} that has been applied to
sound recognition \citep{ellis2011classifying} and synthesis tasks
\citep{mcdermott2011sound}.

The idea of learning from paired audio-visual signals has been studied
extensively in cognitive science \citep{smith2005development}, and
early work introduced computational models for these ideas.
Particularly relevant is the seminal work of de Sa
\citep{de1994learning,de1994minimizing}, which introduced a
self-supervised learning algorithm for jointly training audio and
visual networks.  Their method works on the principle of minimizing
disagreement: they maintain a codebook (represented as a small neural
network) that maps audio and visual examples to a label.  They then
iteratively refine the codebooks until they assign the same labels to
each exemplar.

More recently, researchers have proposed many unsupervised learning
methods that learn visual representations by solving prediction tasks
(sometimes known as {\em pretext} tasks) for which the held-out
prediction target is derived from a natural signal in the world,
rather than from human annotations. This style of learning has been
called {\em self supervision} \citep{de1994minimizing} or ``natural''
supervision~\citep{isola2015thesis}. With these methods, the
supervisory signal may come from video, for example by having the
algorithm estimate camera motion
\citep{agrawal2015learning,jayaraman2015learning} or track content
across frames
\citep{wang2015unsupervised,mobahi2009deep,goroshin2015unsupervised}. There
are also methods that learn from static images, for example by
predicting the relative location of image patches
\citep{doersch2015unsupervised,isola2015learning}, or by learning
invariance to simple geometric and photometric transformations
\citep{dosovitskiy2014discriminative}.  The assumption behind these
methods is that, in order to solve the pretext task, the model will
have to learn about semantics, and therefore through this process
it will learn features that are broadly useful.

While we share with this work the high-level goal of learning image
representations, and we use a similar technical approach, our work
differs in significant ways.  In contrast to methods whose supervisory
signal comes entirely from the imagery itself, ours comes from a
modality (sound) that is complementary to vision. This is advantageous
because sound is known to be a rich source of information about
objects and scenes \citep{gaver1993world,ellis2011classifying}, and
because it is largely invariant to visual transformations, such as
lighting, scene composition, and viewing angle
(\fig{fig:motivation}). Predicting sound from images thus requires
some degree of generalization to visual transformations.  Moreover,
our supervision task is based on solving a straightforward
classification problem, which allows us to use a network design that
closely resembles those used in object and scene recognition (rather
than, for example, the siamese-style networks used in video methods).

Our approach is closely related to recent audio-visual work
\citep{owens2015visually} that predicts soundtracks for videos that
show a person striking objects with a drumstick.  A key feature of
this work is that the sounds are ``visually indicated'' by actions in
video -- a situation that has also been considered in other contexts,
such as in the task of visually localizing a sound source
\citep{hershey1999audio,kidron2005pixels,fisher2000learning} or in
evaluating the synchronization between the two modalities
\citep{slaney2000facesync}.  In the natural videos that we use,
however, the visual motion that produces the sound may not be easily
visible, and sound sources are also frequently out of frame.  Also, in
contrast to other recent work in multi-modal representation learning
\citep{ngiam2011multimodal,srivastava2012multimodal,andrew2013deep},
our technical approach is based on solving a self-supervised
classification problem (rather than fitting a generative model or
autoencoder), and our goal is to learn visual representations that are
generally useful for object recognition tasks.

This work was originally introduced in a conference paper
\citep{owens2016ambient}.  In this expanded version, we include
additional results. In particular: (1) a comparison between an audio
representation learned from unlabeled data and ``ground-truth''
human-annotated audio labels (\sect{sec:role}), (2) additional
visualizations using class activation maps (\sect{sec:vissoundpred}),
and (3) an expanded comparison of our learned image features
(\sect{sec:feateval}).

Since our original publication, researchers have proposed many
interesting audio-visual learning methods.  In particular,
\citet{arandjelovic2017look} solved a similar unsupervised learning
problem, but instead of using hand-crafted audio features, they
jointly learned audio and visual CNNs.  To do this, they used an
embedding-like model, whereby they trained visual and audio CNNs to
predict whether a given pair of audio and visual examples were sampled
from the same video.  Likewise, \citet{aytar2016soundnet} introduced a
method for transferring object labels from visual CNNs to audio CNNs.
To do this, they used a form of cross-modal distillation
\citep{gupta2016cross}, and trained an audio CNN to predict which
semantic labels a pre-trained visual CNN will assign to a paired
audio-visual example.

Researchers have also developed methods for learning from multiple
self-supervision tasks \citep{doersch2017multi}, which could
potentially be used to combined audio-based training with other
methods. They have also developed a variety of new, successful
self-supervised learning approaches, such as methods based on
colorizing grayscale images
\citep{zhang2016colorful,zhang2016splitbrain} and predicting how
objects move \citep{pathak2016learning}.  We include additional
comparisons with these new methods.

There has also been recent work in visualizing the internal
representation of a neural net.  For example, \citet{bau2017network}
quantified the number of object-selective units in our network, as
well as other recent networks trained with self-supervised learning.
This work arrived at a similar conclusion as in this work, with a
different visualization methodology: namely, that the model learned
units that are selective to objects.

\makeatletter{}%
\begin{figure*}[t!]
  \includegraphics[width=0.98\linewidth]{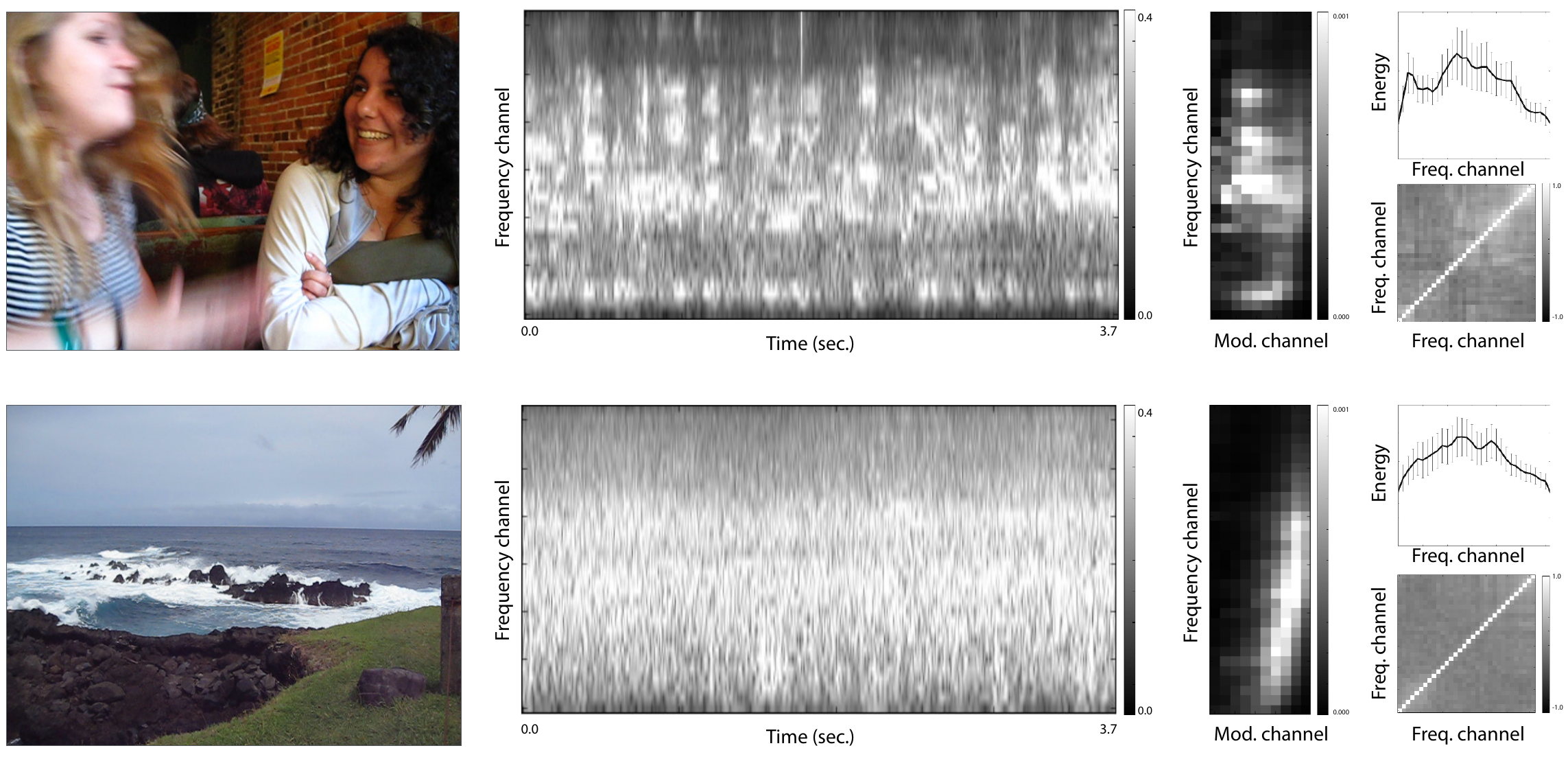}
  {\small \noindent \phantom{~~~~~~~~~~~~~~~~~~~~~} (a) Video frame \hspace{45mm} (b) Cochleagram \hspace{35mm} (c) Summary statistics}
  \caption{Visual scenes are associated with characteristic
    sounds. Our goal is to take an image (a) and predict time-averaged
    summary statistics (c) of a cochleagram (b).  The statistics we
    use are (clockwise): the response to a bank of band-pass
    modulation filters (sorted left-to-right in increasing order of
    frequency); the mean and standard deviation of each frequency
    band; and the correlation between bands.  We show two frames from
    the YFCC100m dataset \citep{thomee2015yfcc100m}.  The first
    contains the sound of human speech; the second contains the sound
    of wind and crashing waves.  The differences between these sounds
    are reflected in their summary statistics: \eg, the water/wind
    sound, which is similar to white noise, contains fewer
    correlations between cochlear channels. }
  \label{fig:soundtex}
\end{figure*}
\makeatletter{}%
\makeatletter{}%
\begin{figure*}[t!]

\includegraphics[width=\linewidth]{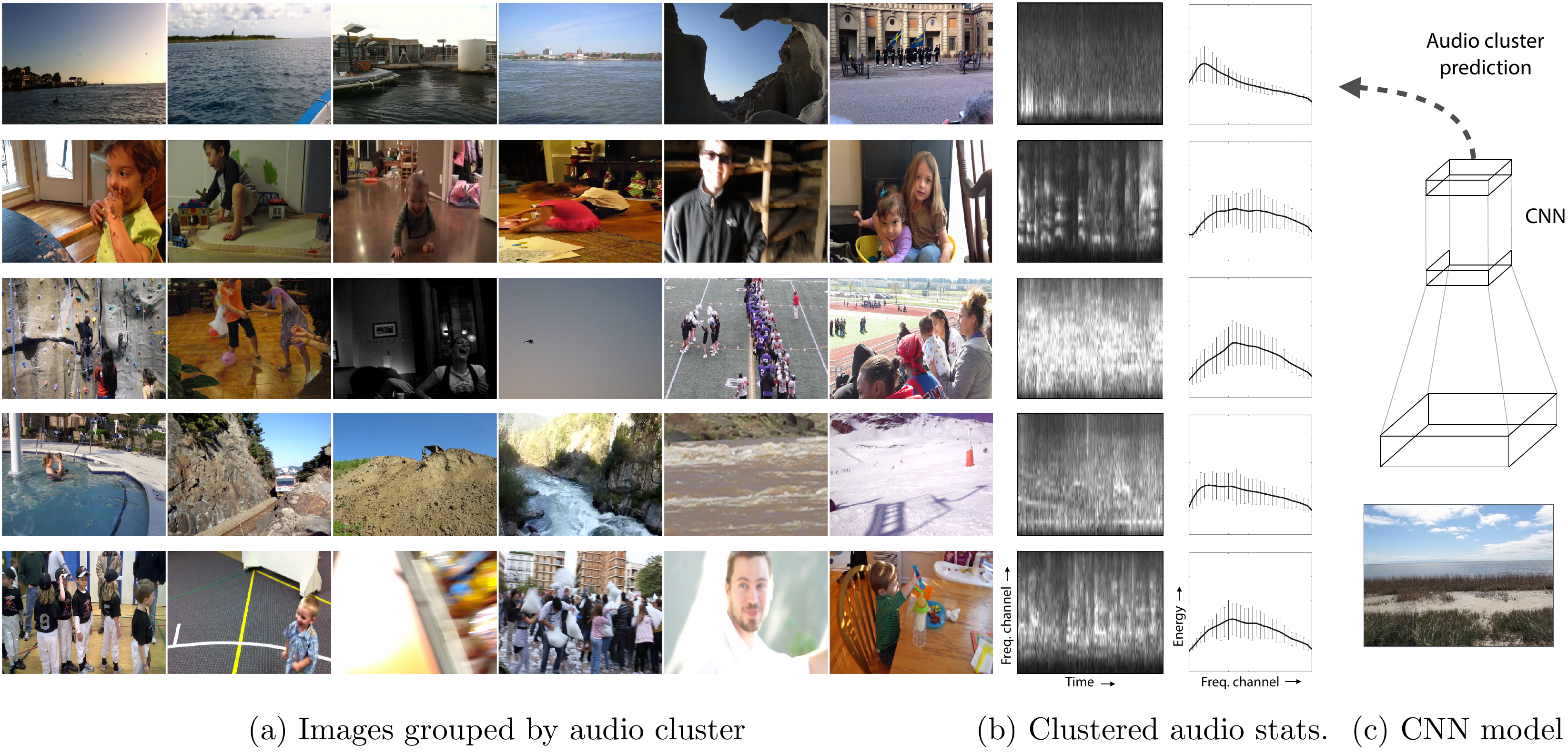}
\caption{Visualization of some of the audio clusters used in our model (5 of 30 clusters).  For each cluster, we show (a) the images in the test set whose sound textures were closest to the centroid (no more than one frame per video), and (b) we visualize aspects of the sound texture used to define the cluster centroid -- specifically, the mean and standard deviation of the frequency channels.  We also include a representative cochleagram  (that of the leftmost image).  Although the clusters were defined using audio, there are common objects and scene attributes in many of the images.  We train a CNN to predict a video frame's auditory cluster assignment (c).}
\label{fig:soundclusters}
  
\end{figure*}

\section{Learning to predict ambient audio}

\label{sec:predict}

We would like to train a model that, when given a frame of video, can predict its corresponding sound -- a task that requires knowledge of objects and scenes, among other factors like human behavior, semantics, and culture.

\subsection{Statistical sound summaries} A natural question, then, is how our model should represent sound. Perhaps the first approach that comes to mind would be to estimate a frequency spectrum at the moment in which the picture was taken, similar to \cite{owens2015visually}.  However, this is potentially suboptimal because in natural scenes it is difficult to predict the precise timing of a sound from visual information.  Upon seeing a crowd of people, for instance, we might expect to hear the sound of speech, but the precise timing and content of that speech might not be directly indicated by the video frames.  

To be closer to the time scale of visual objects, we estimate a
statistical summary of the sound, averaged over a few seconds of
audio.  While there are many possible audio features that could be
used to compute this summary, we use the perceptually inspired sound
texture model of \cite{mcdermott2011sound}, which assumes that the
audio is stationary within a temporal window (we use \winsizesec
seconds).  More specifically, we closely
follow \cite{mcdermott2011sound} and filter the audio waveform with a
bank of \numfullbands band-pass filters intended to mimic human
cochlear frequency selectivity (producing a representation similar to
a spectrogram). We then take the Hilbert envelope of each channel,
raise each sample of the envelope to the 0.3 power (to mimic cochlear
amplitude compression), and resample the compressed envelope to 400
Hz. Finally, we compute time-averaged statistics of these subband
envelopes: we take the mean and standard deviation of each
frequency channel, the mean squared response of each of a bank of
modulation filters applied to each channel, and the Pearson
correlation between pairs of channels. For the modulation filters, we
use a bank of 10 band-pass filters with center frequencies ranging
from 0.5 to 200 Hz, equally spaced on a logarithmic scale.

To make the sound features more invariant to gain (e.g., from the microphone), we divide the envelopes by the median energy (median vector norm) over all timesteps, and include this energy as a feature. As in \cite{mcdermott2011sound}, we normalize the standard deviation of each cochlear channel by its mean, and each modulation power by its standard deviation.  We then rescale each kind of texture feature (\ie marginal moments, correlations, modulation power, energy) inversely with the number of dimensions. The sound texture for each image is a \soundtexdim-dimensional vector. In \fig{fig:soundtex}, we give examples of these summary statistics for two audio clips.  We provide more details about our audio representation in \sect{sec:soundtex}.

\subsection{Predicting ambient sound from images}
\label{sec:cluster}
We would like to predict sound textures from images -- a task that we hypothesize leads to learning useful visual representations.  Although multiple frames are available, we predict sound from a single frame, so that the learned image features will be more likely to transfer to single-image recognition tasks.  Furthermore, since the actions that produce the sounds may not appear on screen, motion information may not always be applicable.

While one option would be to regress the sound texture $v_j$ directly from the corresponding image $I_j$, we choose instead to define explicit sound categories and formulate this visual recognition problem as a classification task. This also makes it easier to analyze the network, because it allows us to compare the internal representation of our model to object- and scene-classification models with similar network architecture (\sect{sec:objdet}). We consider two labeling models: one based on a vector quantization, the other based on a binary coding scheme.

\vpar{Clustering audio features} In the {\em \clustermod} model, the sound textures $\{v_j\}$ in the training set are clustered using $k$-means.  These clusters define image categories: we label each sound texture with the index of the closest centroid, and train our CNN to label images with their corresponding labels.

We found that audio clips that belong to a cluster often contain common objects.  In \fig{fig:soundclusters}, we show examples of such clusters, and in the supplementary material we provide their corresponding audio.  We can see that there is a cluster that contains indoor scenes with children in them; these are relatively quiet scenes punctuated with speech sounds. Another cluster contains the sounds of many people speaking at once (often large crowds); another contains many water scenes (usually containing loud wind sounds).  Several clusters capture general scene attributes, such as outdoor scenes with light wind sounds.  During training, we remove examples that are far from the centroid of their cluster (more than the median distance to the vector, amongst all examples in the dataset).

\vpar{Binary coding model} For the other variation of our model (which we call the {\em \binmod} model), we use a binary coding scheme~\citep{indyk1998approximate,salakhutdinov2009semantic,weiss2009spectral} equivalent to a multi-label classification problem.  We project each sound texture $v_j$ onto the top principal components (we use 30 projections), and convert these projections into a binary code by thresholding them.  We predict this binary code using a sigmoid layer, and during training we measure error using cross-entropy loss.

For comparison, we also trained a model (which we call the {\em \spectmodel} model) to approximately predict the frequency spectrum at the time that the photo was taken, in lieu of a full sound texture.  Specifically, for our sound vectors $v_j$ in this model, we used the mean value of each cochlear channel within a 33.3 ms interval centered on the input frame (approximately one frame of a 30 Hz video).  For training, we used the projection scheme from the \binmod model.

\vpar{Training} We trained our models on a 360,000-video subset of the YFCC100m video dataset~\citep{thomee2015yfcc100m} (which we also call the {\em Flickr video} dataset).  A large fraction of the videos in the dataset are personal video recordings containing natural audio, though many were post-processed, \eg with added subtitles, title screens, and music.  We divided our videos into training and test sets, and we randomly sampled \numpervideo frames per video (\numtrainimsmil million training images total). For our network architecture, we used the CaffeNet architecture~\citep{jia2014caffe}, a variation of \cite{krizhevsky2012imagenet}, with batch normalization~\citep{ioffe2015batch}.  We trained our model with Caffe~\citep{jia2014caffe}, using a batch size of 256, for 320,000 iterations of stochastic gradient descent with momentum, decreasing the learning rate from an initial value of 0.01 by a factor of 10 every 100,000 iterations.

\makeatletter{}%

\makeatletter{}%

\begin{figure}[t!]
  \centering
  \includegraphics[width=0.98\linewidth]{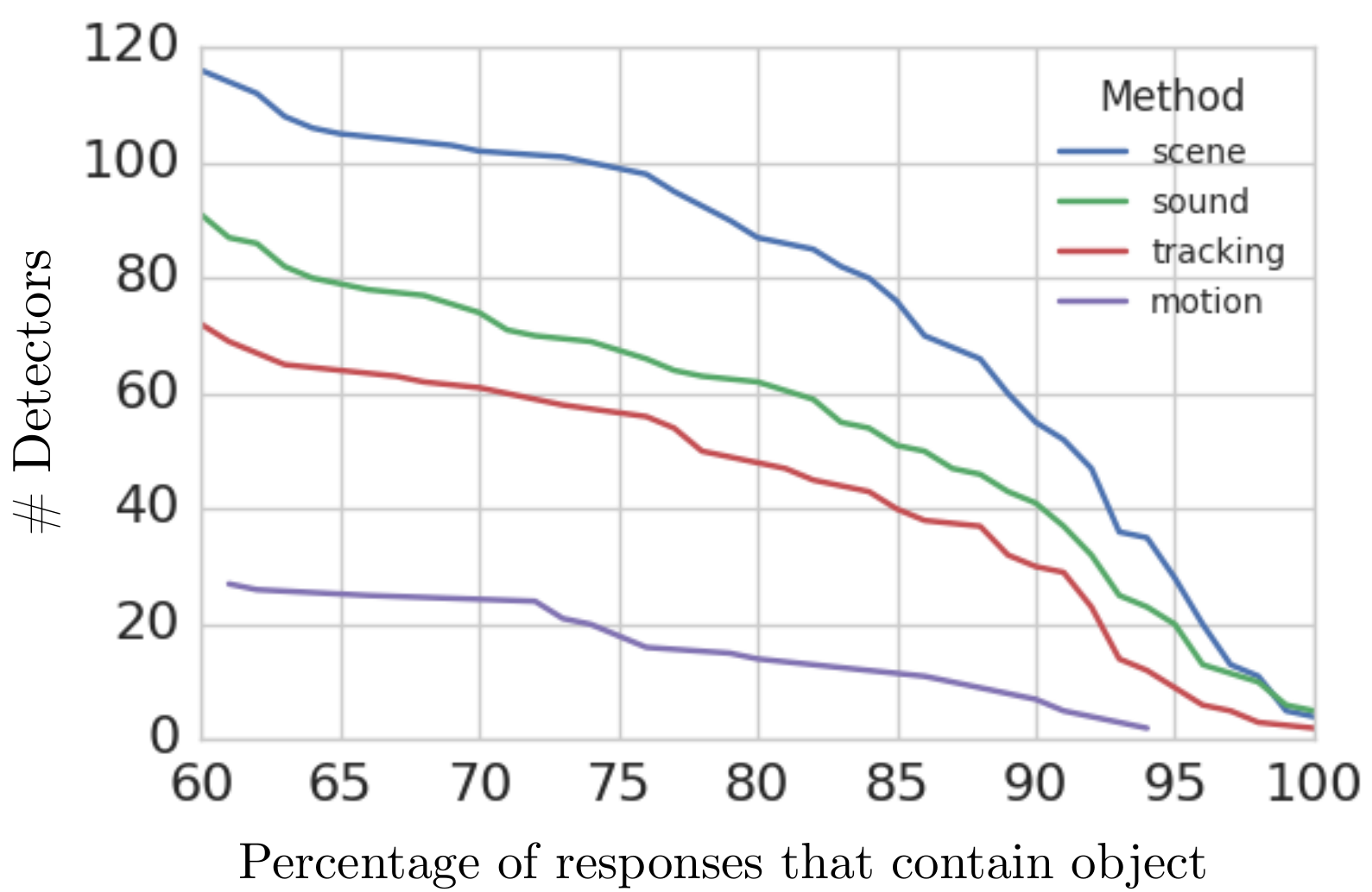}
    \caption{The number of object-selective units for each method,
  as we increase the threshold used to determine whether a unit is
  object-selective.  This threshold corresponds to the fraction of
  images that contain the object in question, amongst the images with
  the 60 largest activations.  For our analysis in \sect{sec:results},
  we used a threshold of 60\%.} \label{fig:threshold-curve}
\end{figure}
\makeatletter{}%
\begin{figure*}[t!]
\centering
\small
\begin{tabular}{ccc}
     \multicolumn{3}{c}{Neuron visualizations of the network trained by \bf sound} \\
     field & sky & grass \\
     \includegraphics[width=0.3\linewidth]{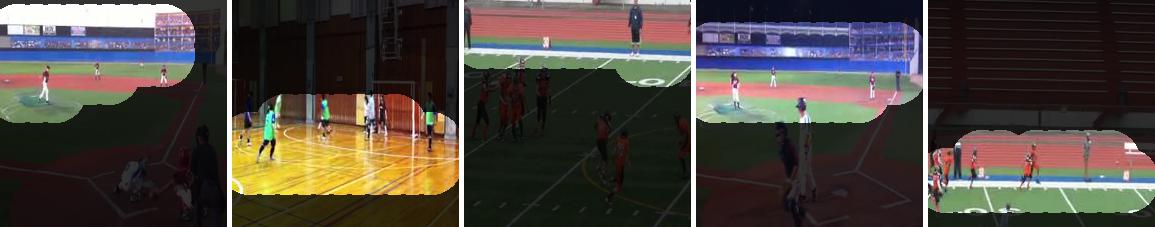} 
     & \includegraphics[width=0.3\linewidth]{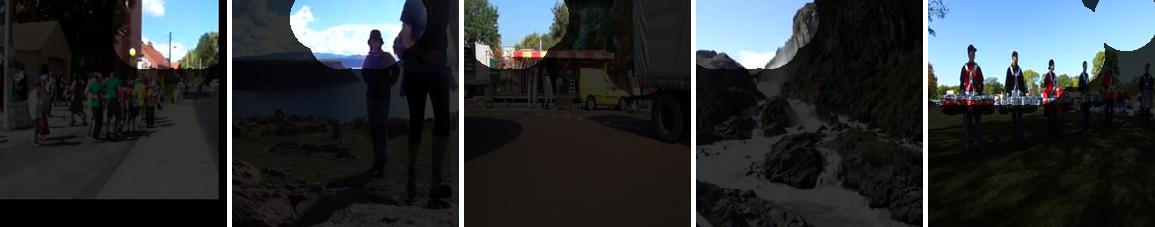} 
     & \includegraphics[width=0.3\linewidth]{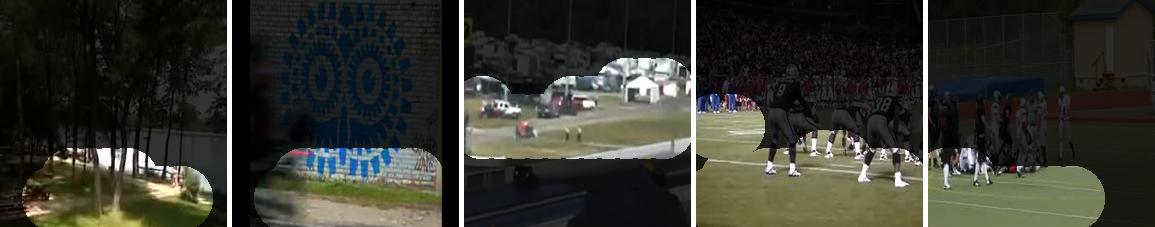} \\
     snowy ground & ceiling & car\\
     \includegraphics[width=0.3\linewidth]{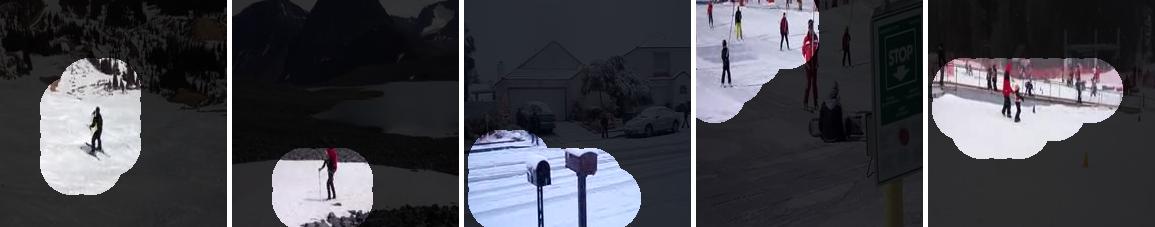}
     & \includegraphics[width=0.3\linewidth]{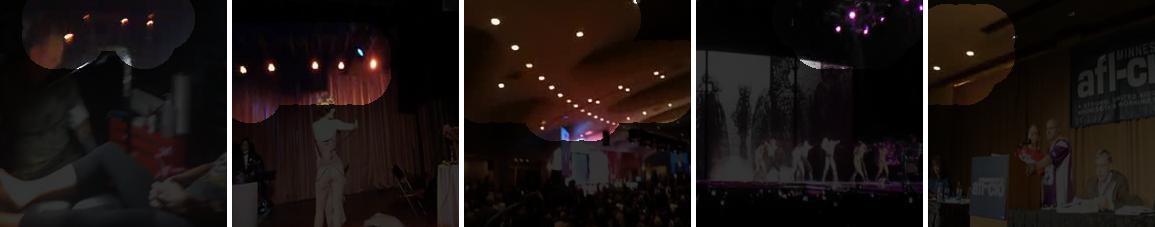} 
     & \includegraphics[width=0.3\linewidth]{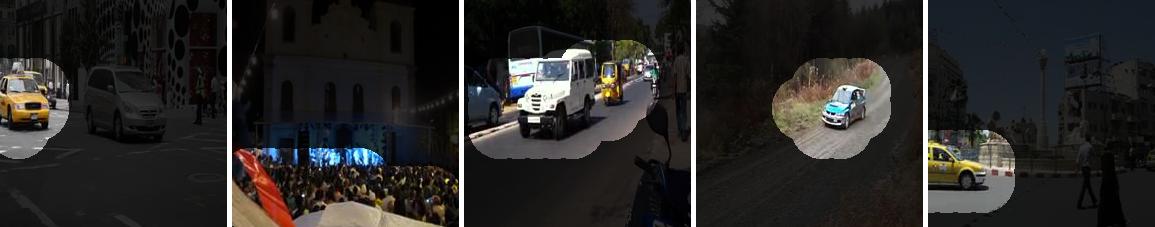} \\
     waterfall & waterfall & sea \\
     \includegraphics[width=0.3\linewidth]{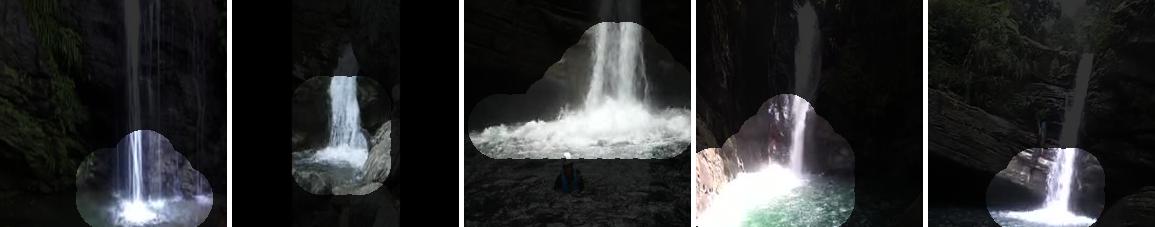}
     & \includegraphics[width=0.3\linewidth]{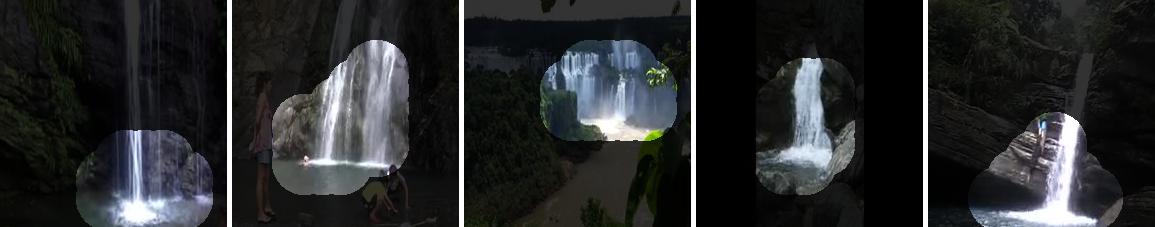}
     & \includegraphics[width=0.3\linewidth]{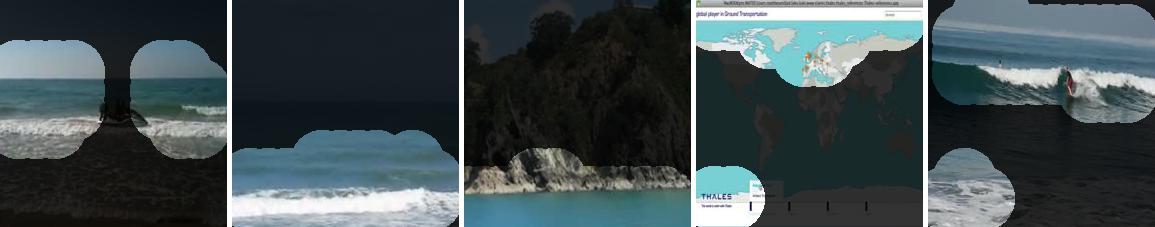} \\
      baby &  baby &  baby\\
     \includegraphics[width=0.3\linewidth]{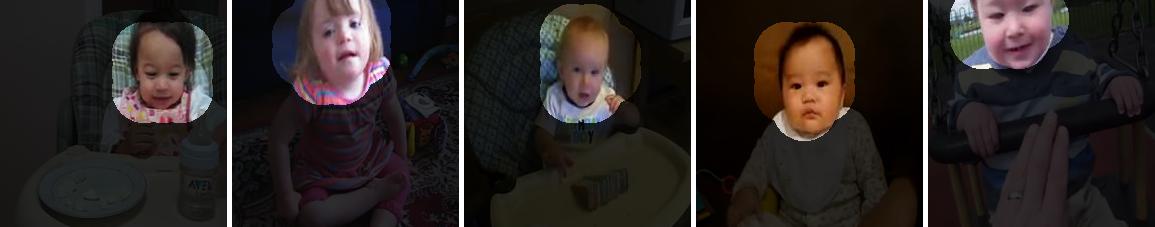}
     & \includegraphics[width=0.3\linewidth]{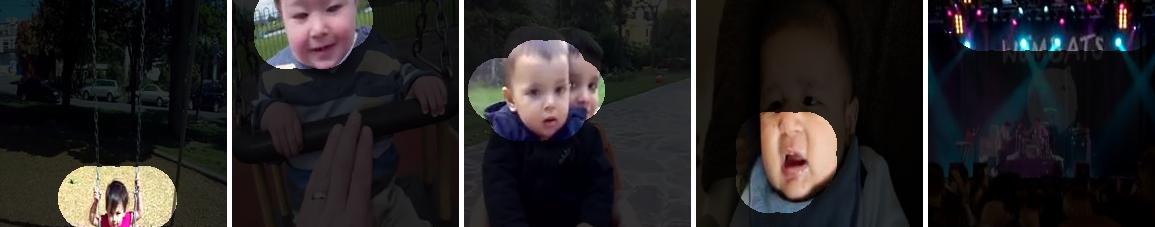}
     & \includegraphics[width=0.3\linewidth]{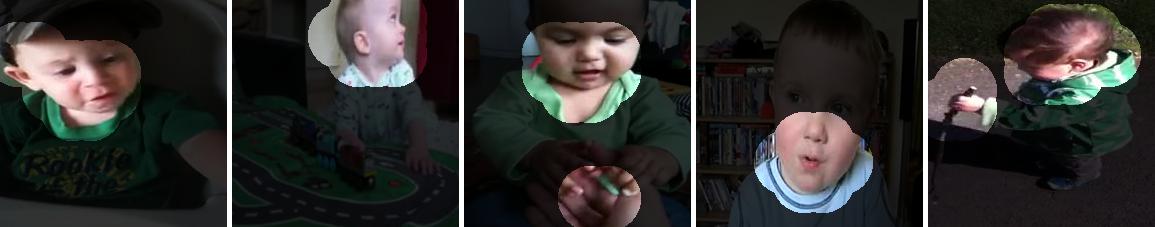} \\
      person &  person &  person \\
     \includegraphics[width=0.3\linewidth]{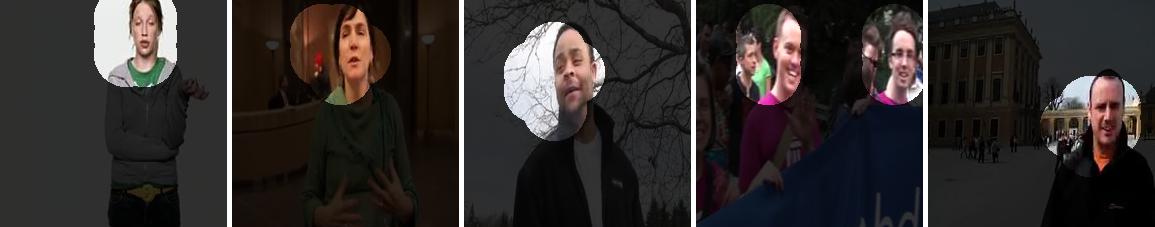}
     & \includegraphics[width=0.3\linewidth]{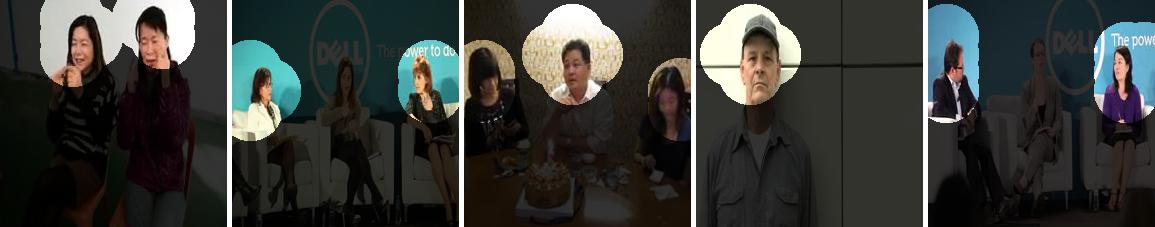}
     & \includegraphics[width=0.3\linewidth]{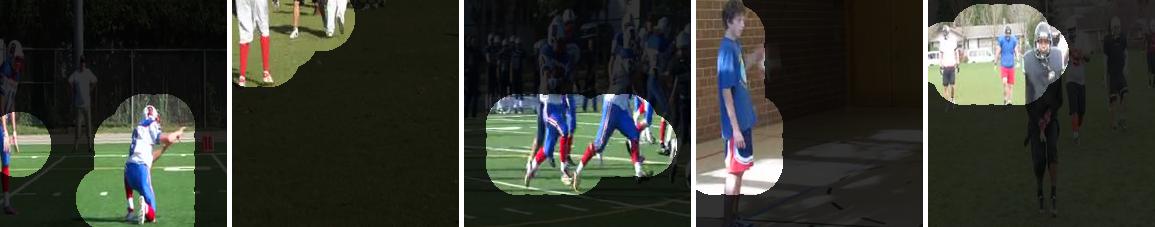} \\
       person &  person &  person  \\
     \includegraphics[width=0.3\linewidth]{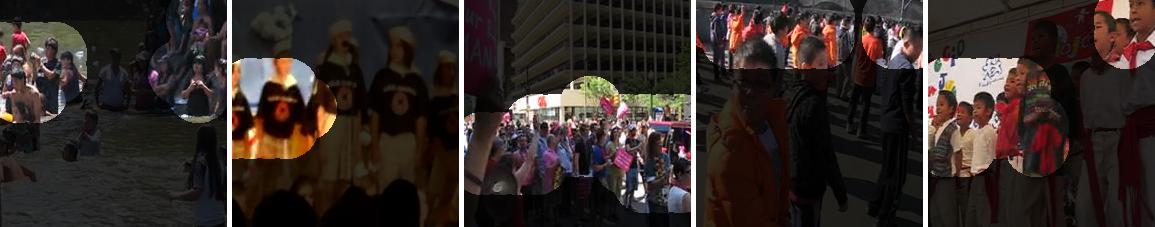}
     & \includegraphics[width=0.3\linewidth]{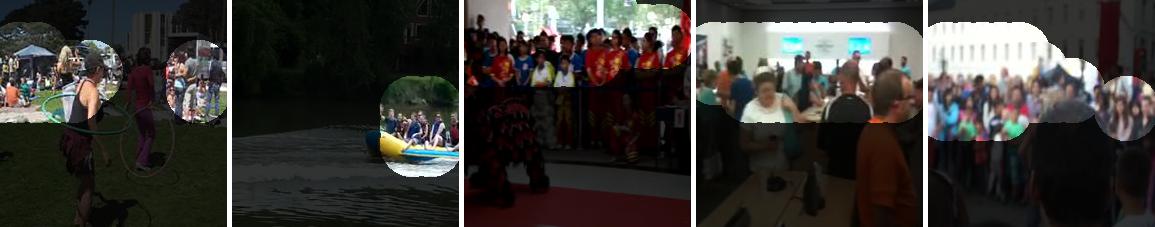}
     & \includegraphics[width=0.3\linewidth]{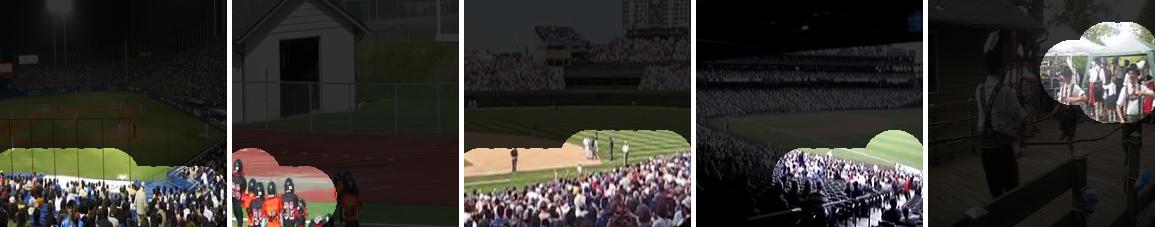} \\
      grandstand &  grandstand &  grandstand \\
     \includegraphics[width=0.3\linewidth]{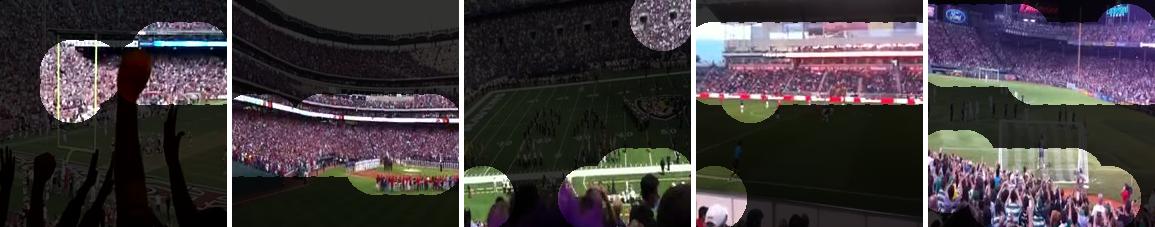}
     & \includegraphics[width=0.3\linewidth]{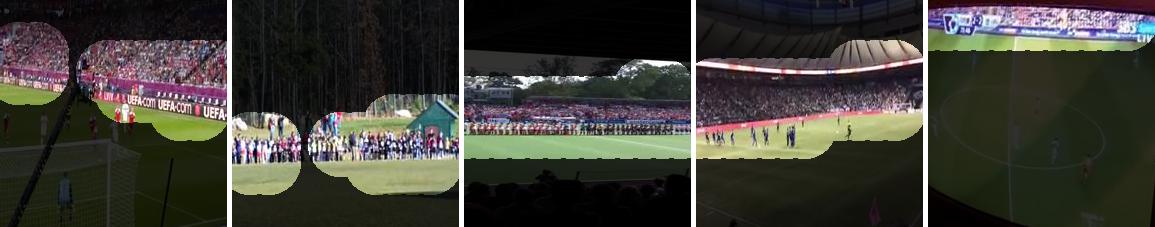}
     & \includegraphics[width=0.3\linewidth]{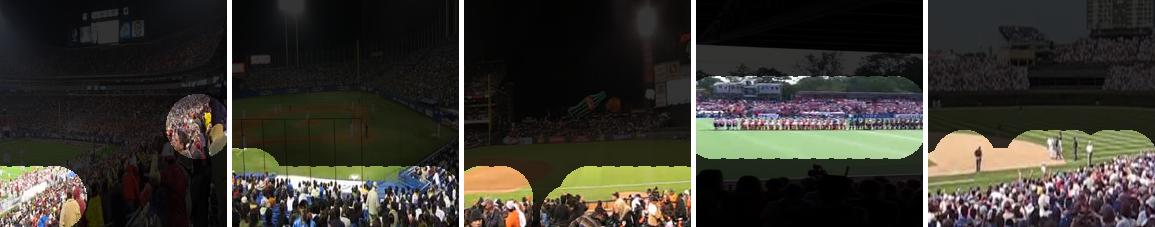} \\
     \midrule
     \multicolumn{3}{c}{Neuron visualizations of the network trained by visual {\bf tracking} \citep{wang2015unsupervised}} \\
      sea &  grass &  road \\
     \includegraphics[width=0.3\linewidth]{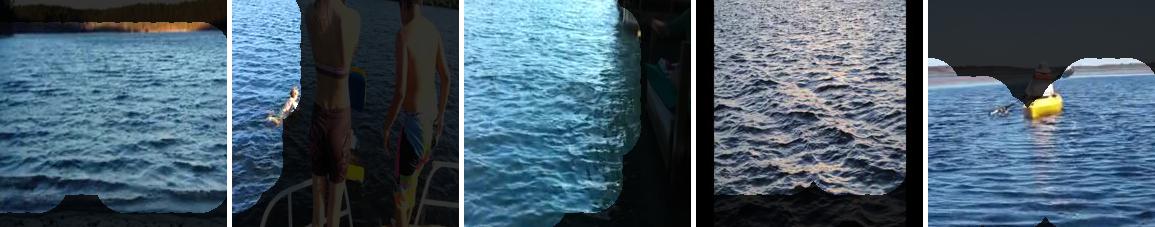} 
     & \includegraphics[width=0.3\linewidth]{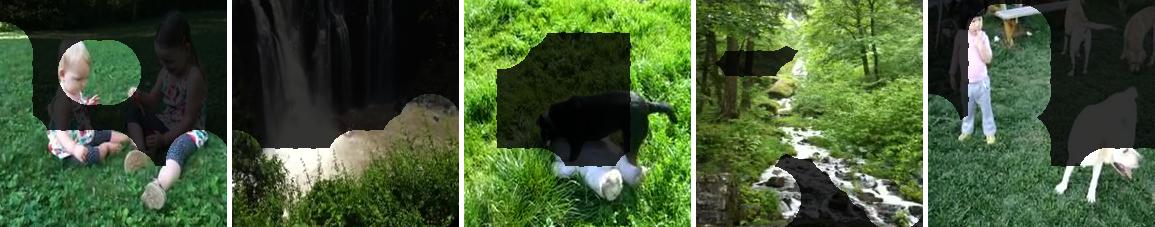} 
     & \includegraphics[width=0.3\linewidth]{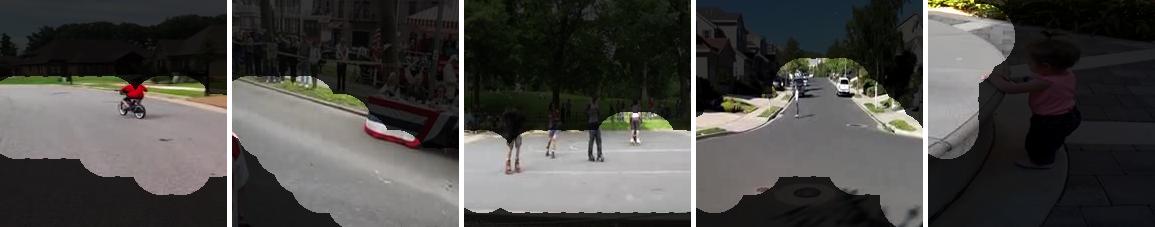} \\
      sea &  pitch &  sky \\
     \includegraphics[width=0.3\linewidth]{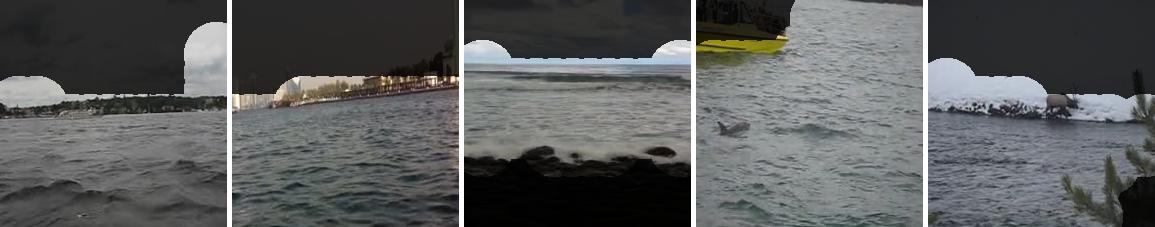}
     & \includegraphics[width=0.3\linewidth]{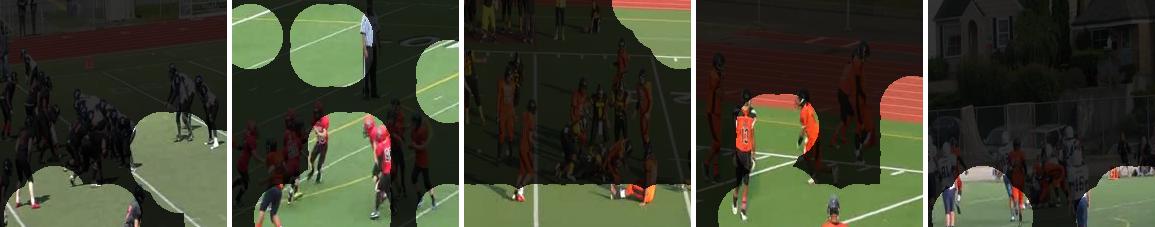}
     & \includegraphics[width=0.3\linewidth]{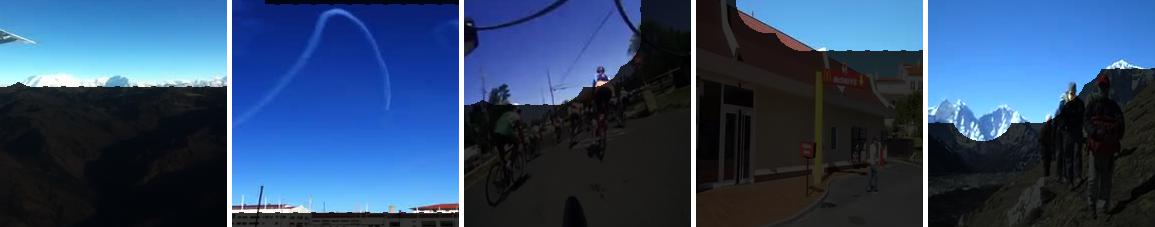} \\
     \midrule
     \multicolumn{3}{c}{Neuron visualizations of the network trained by {\bf egomotion} \citep{agrawal2015learning}} \\
      ground &  sky &  grass\\
     \includegraphics[width=0.3\linewidth]{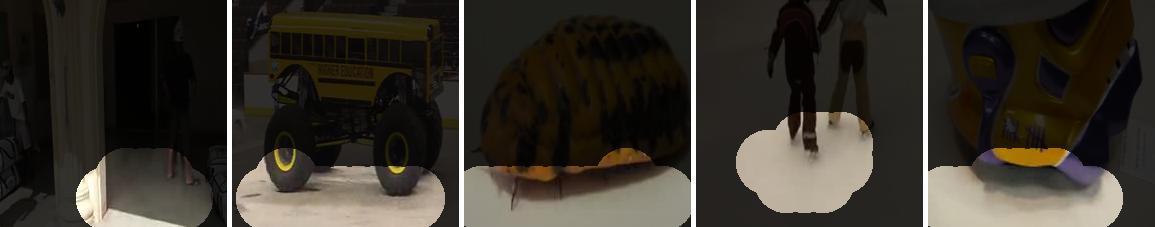} 
     & \includegraphics[width=0.3\linewidth]{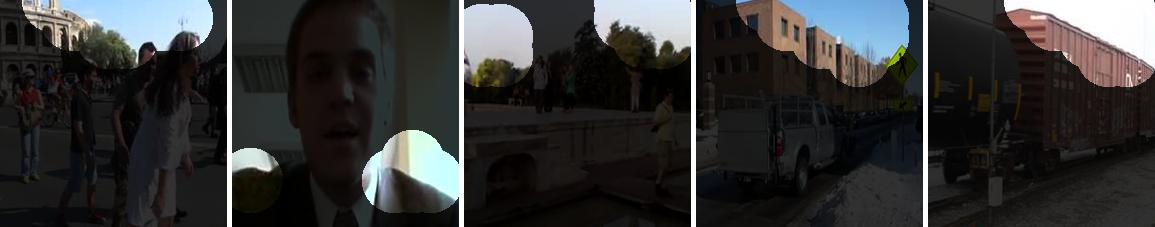} 
     & \includegraphics[width=0.3\linewidth]{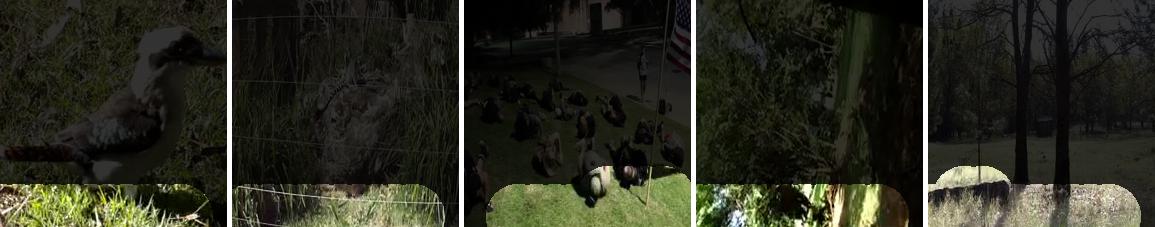} \\
      ground &  sky &  plant\\
     \includegraphics[width=0.3\linewidth]{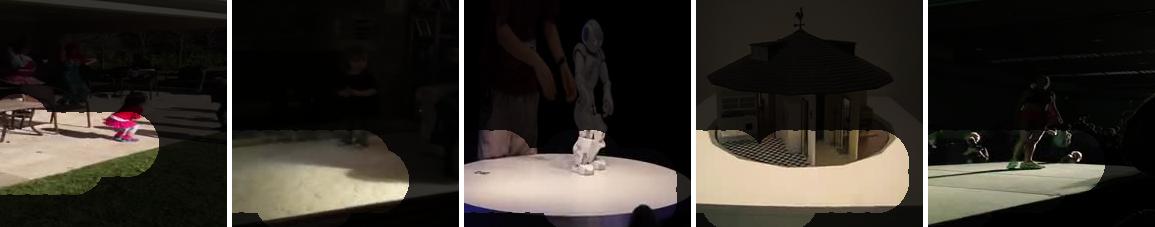}
     & \includegraphics[width=0.3\linewidth]{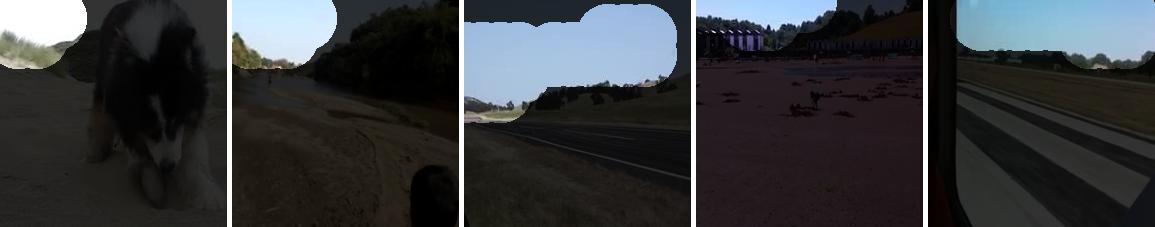}
     & \includegraphics[width=0.3\linewidth]{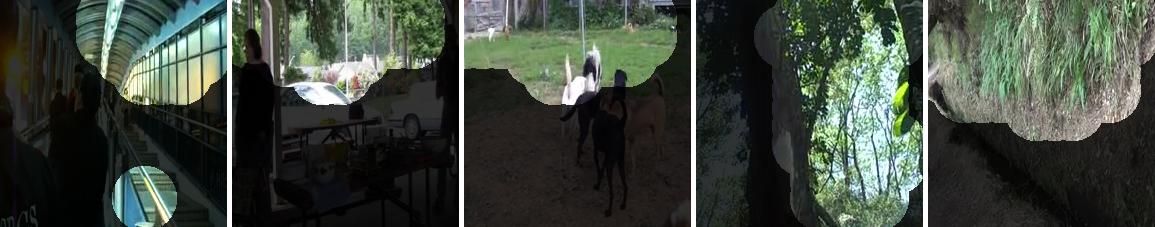} \\
     \midrule
     \multicolumn{3}{c}{Neuron visualizations of the network trained by {\bf patch positions} \citep{doersch2015unsupervised}} \\
      sky &  sky &  baby\\
     \includegraphics[width=0.3\linewidth]{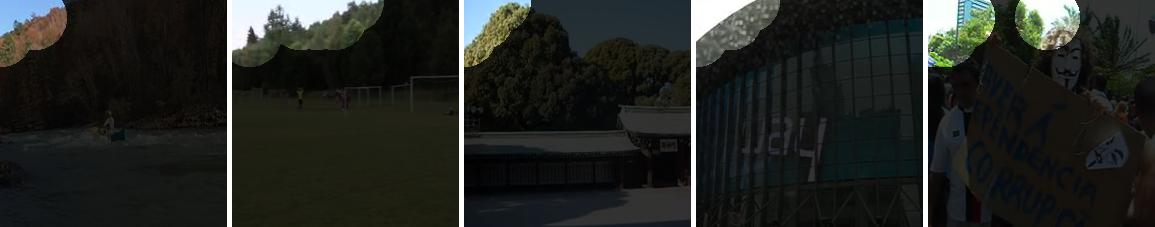}
     & \includegraphics[width=0.3\linewidth]{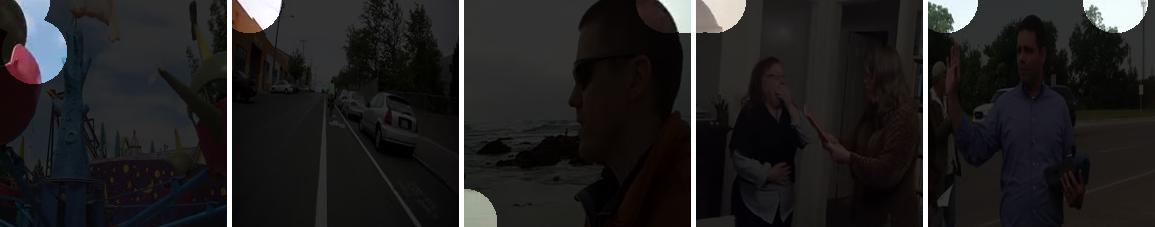}
     & \includegraphics[width=0.3\linewidth]{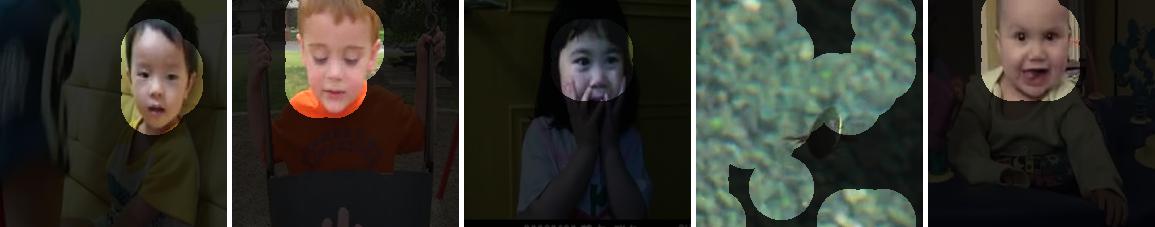} \\
     \midrule
     \multicolumn{3}{c}{Neuron visualizations of the network trained by labeled {\bf scenes} \citep{zhou2014places}} \\
      field &  tent &  building\\
     \includegraphics[width=0.3\linewidth]{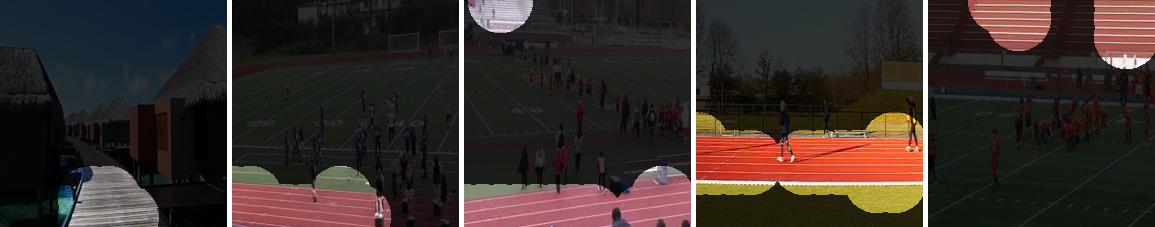} 
     & \includegraphics[width=0.3\linewidth]{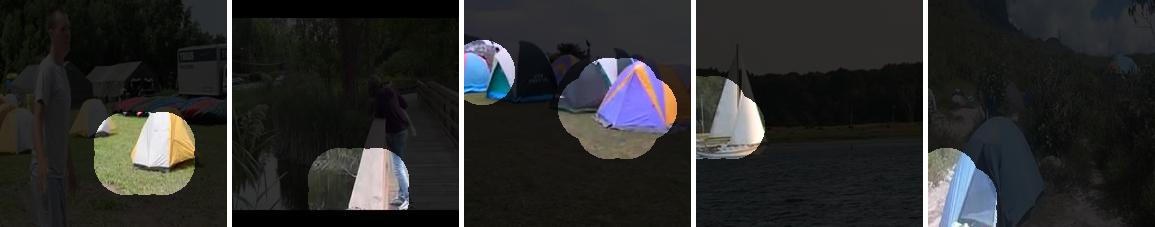} 
     & \includegraphics[width=0.3\linewidth]{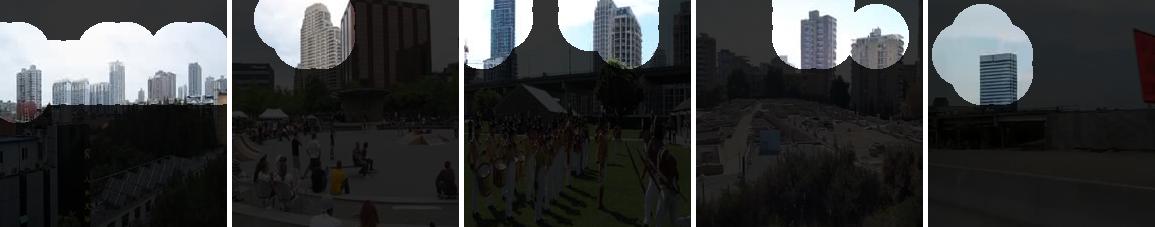} \\
      pitch & path &  sky\\
     \includegraphics[width=0.3\linewidth]{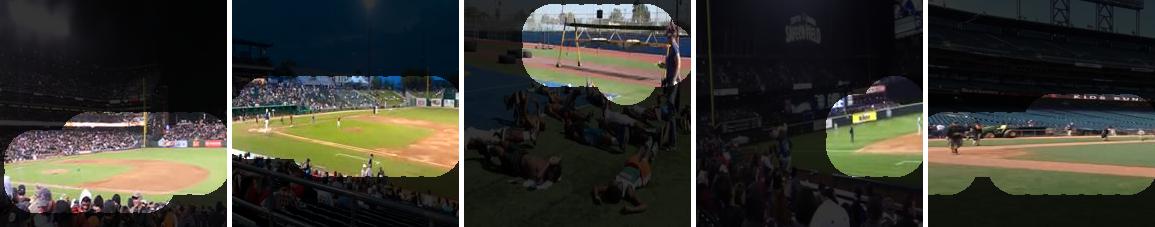} 
     & \includegraphics[width=0.3\linewidth]{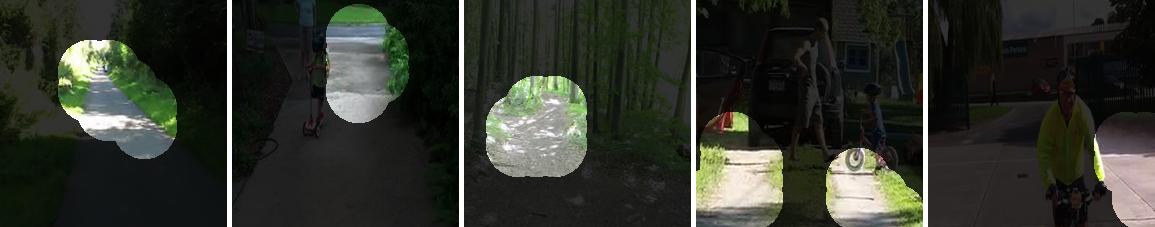} 
     & \includegraphics[width=0.3\linewidth]{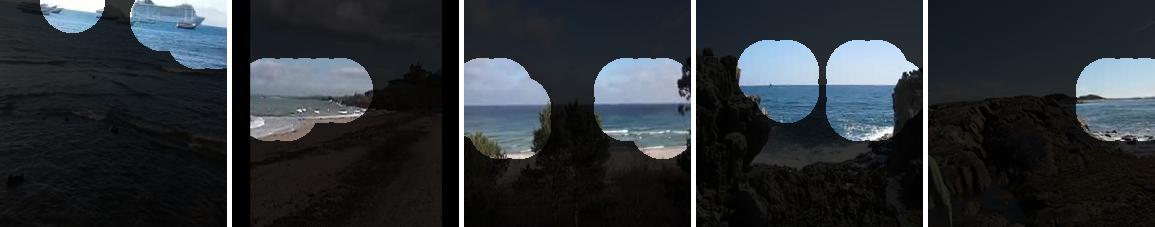}
\end{tabular}
\vspace{-5pt}
  \caption{Top 5 responses for convolutional units in various networks, evaluated on videos from the YFCC100m dataset.}
  \label{fig:neuronsamples}
  \vspace{-20pt}
\end{figure*}

\makeatletter{}%
\begin{figure*}[t!]
    \centering
  \vspace{0.5mm}Training by sound (\numobjunits Detectors) \\
  \vspace{-0.4mm}
  \includegraphics[width=\linewidth]{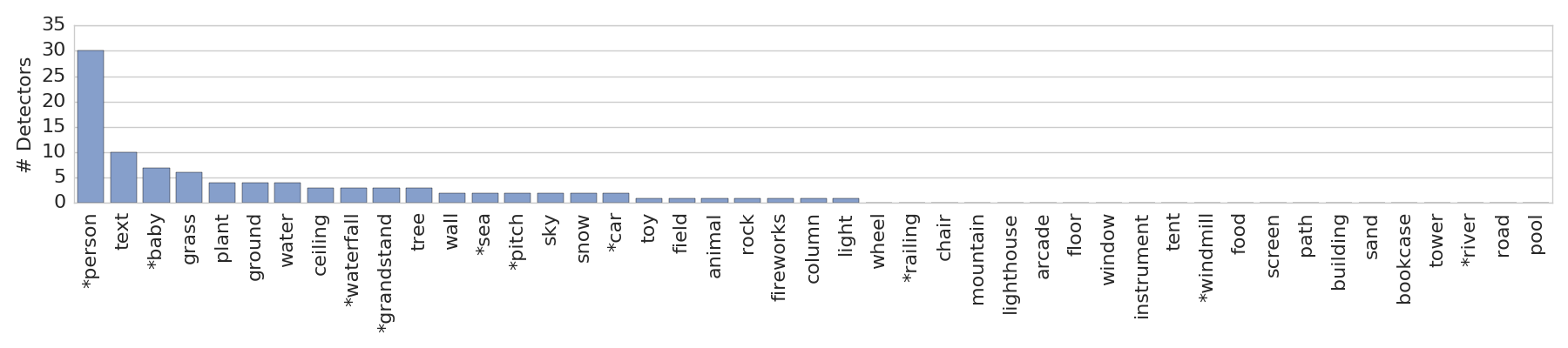} \\
  \ \\
  Training by labeled scenes \citep{zhou2014places} (\numplacesunits Detectors) \\
  \vspace{-0.4mm}
  \includegraphics[width=\linewidth]{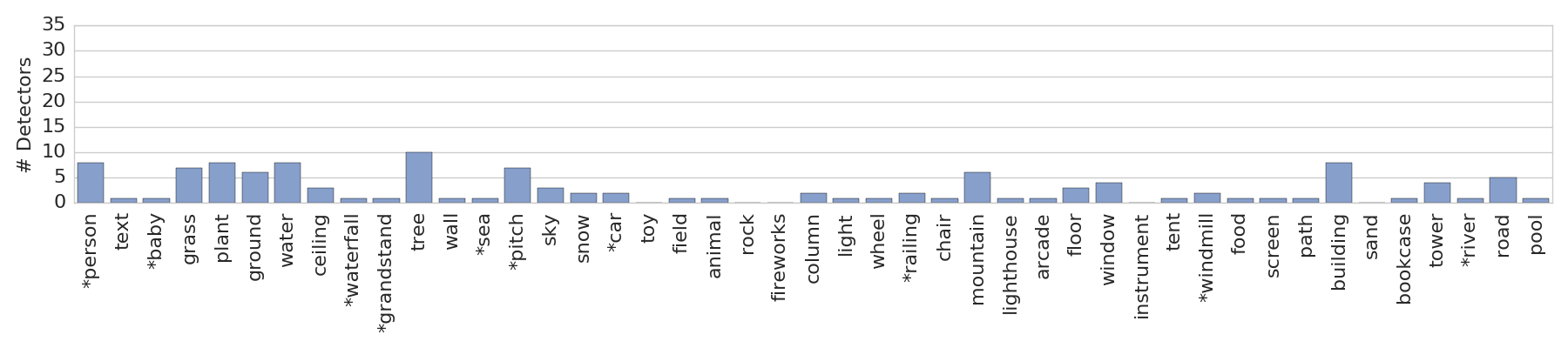} \\
  \ \\
  Training by visual tracking \citep{wang2015unsupervised} (\numtrackingunits Detectors)\\
  \vspace{-0.4mm}
  \includegraphics[width=\linewidth]{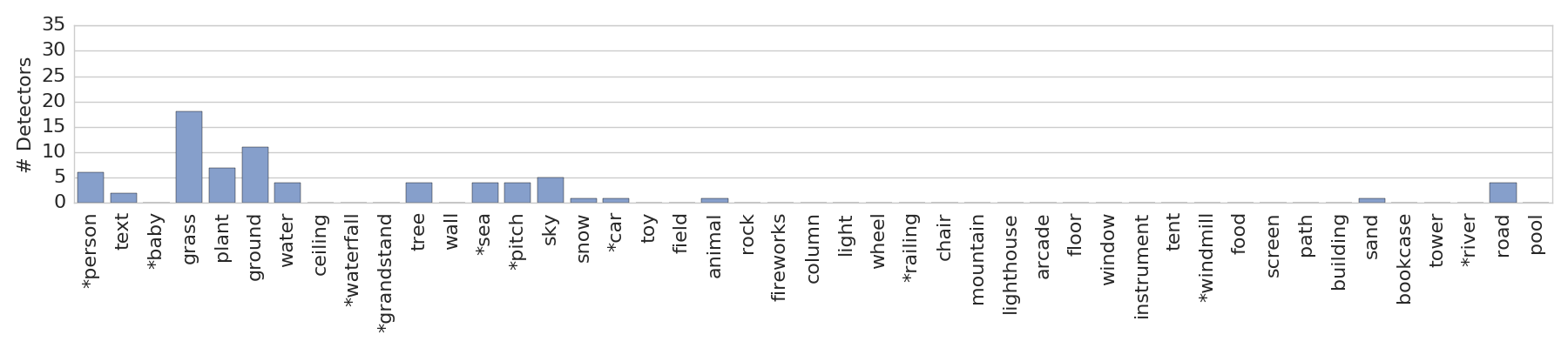}
  \caption{Histogram of object-selective units in networks trained with different styles of supervision. From top to bottom: training to predict ambient sound (our Clustering model); training to predict scene category using the Places dataset \citep{zhou2014places}; and training to do visual tracking \citep{wang2015unsupervised}. Compared to the tracking model, which was also trained without semantic labels, our network learns more high-level object detectors. It also has more detectors for objects that make characteristic sounds, such as {\em person}, {\em baby}, and {\em waterfall}, in comparison to the one trained on Places. Categories marked with $*$ are those that we consider to make characteristic sounds.}
  \label{fig:objdistr}
\end{figure*}
\makeatletter{}%
\begin{figure*}[t!]
    \centering
  Training by sound (67 detectors)\\
  \includegraphics[width=\linewidth]{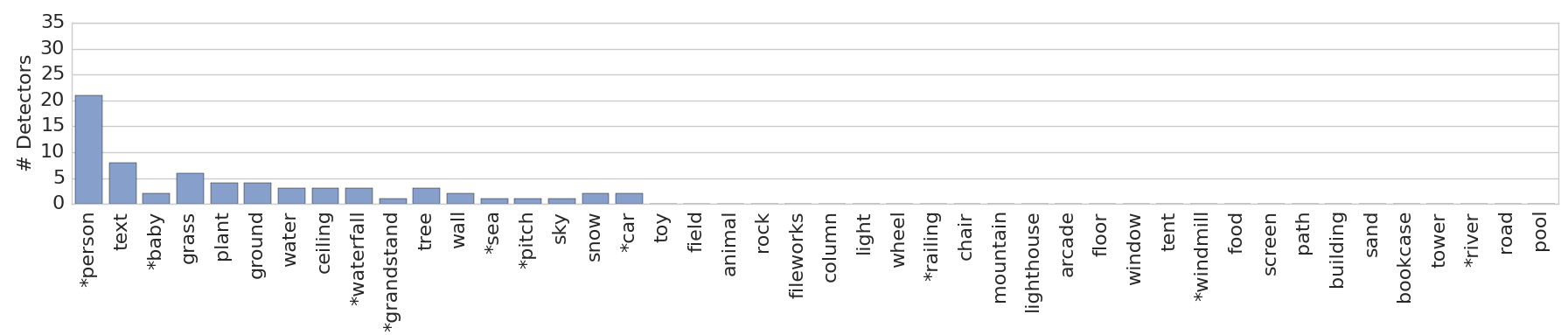} \\
  \ \\
  Training by labeled scenes \citep{zhou2014places} (146 detectors)\\
  \includegraphics[width=\linewidth]{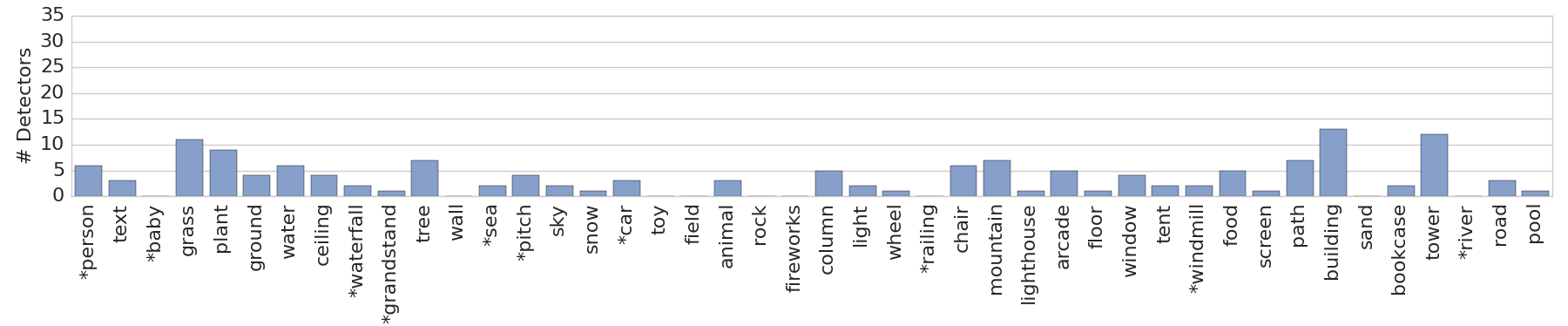} \\
  \ \\
  Training by visual tracking \citep{wang2015unsupervised} (61 detectors)\\
  \includegraphics[width=\linewidth]{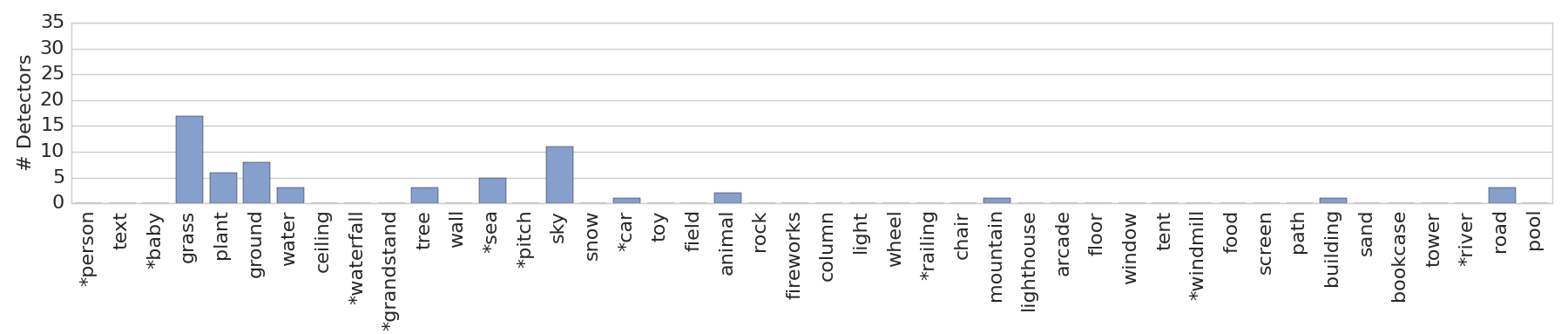}\\
  \caption{The number of object-selective per category, when
    evaluating the model on the SUN and ImageNet datasets
    (cf. \fig{fig:objdistr}, in which the models were evaluated on the
    YFCC100m video dataset).}
  \label{fig:objsun}
\end{figure*}
\makeatletter{}%
\begin{figure*}[t!]
\scriptsize\centering
\begin{tabular}{ccc}
person&car&ceiling\\
\includegraphics[width=0.31\linewidth]{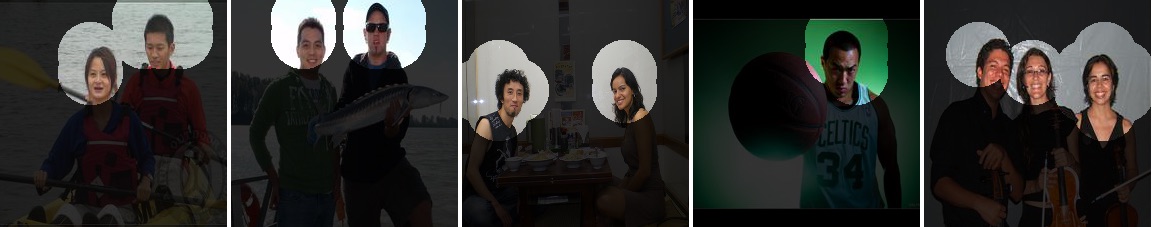}
&\includegraphics[width=0.31\linewidth]{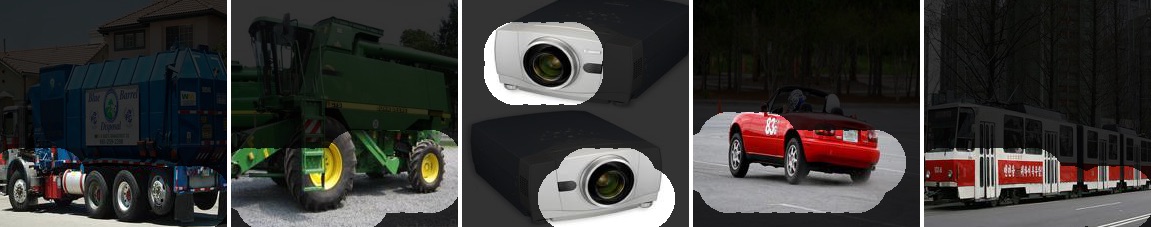}
&\includegraphics[width=0.31\linewidth]{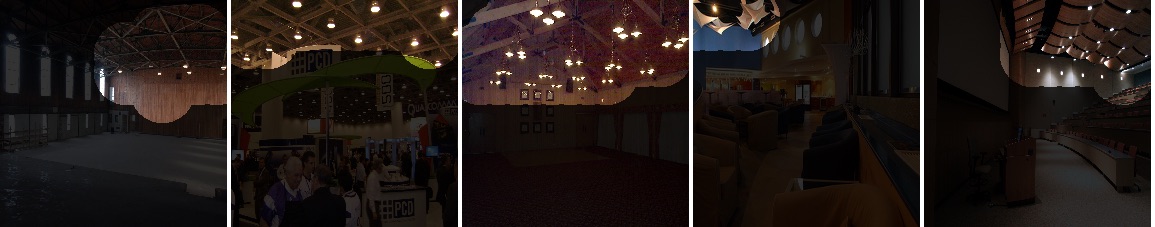}
\\
waterfall%
&text&pitch\\
\includegraphics[width=0.31\linewidth]{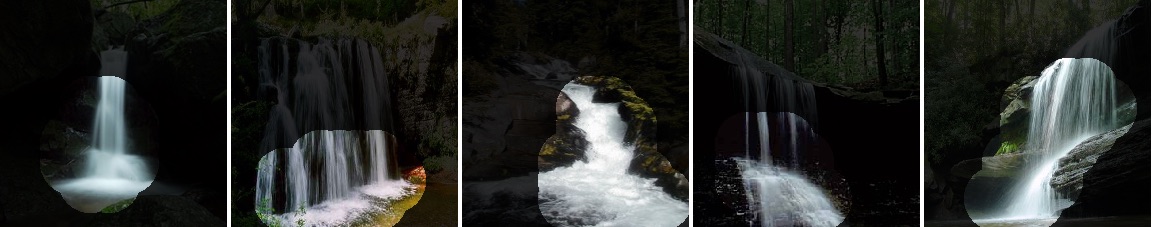}
&\includegraphics[width=0.31\linewidth]{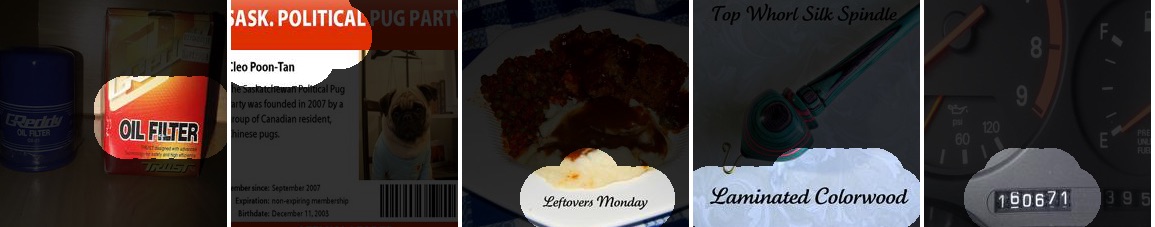}
&\includegraphics[width=0.31\linewidth]{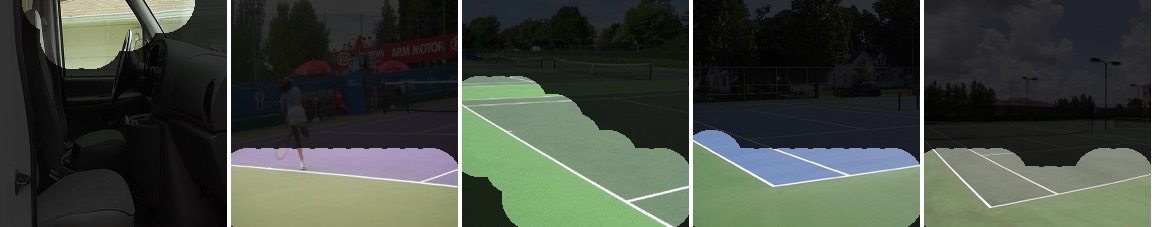}
\\
\end{tabular}
\caption{A selection of object-selective neurons, obtained by testing our model on the SUN and ImageNet datasets. We show the top 5 activations for each unit.}
\label{fig:sun-neuron-vis}
\end{figure*}
\makeatletter{}%

\begin{table*}[t!]
    \centering
  \begin{tabular}{lC{1.5cm}C{1.5cm}}
  \toprule
  Method & Sound & Places \\
  \midrule
       \# Detectors & \numobjunits & \numplacesunits \\
       \# Detectors for objects with characteristic sounds & 49 & 26\\
       Videos with object sound & \fracvidsound & 16.9\% \\
       Characteristic sound rate & 81.2\% & 75.9\%\\
    \bottomrule
  \end{tabular}
  \vspace{5pt}
  \caption{Row 1: the number of detectors (\ie units that are selective to a particular object); row 2: the number of detectors for objects with characteristic sounds; row 3: fraction of videos in which an object's sound is audible (computed only for object classes with characteristic sounds); row 4: given that an activation corresponds to an object with a characteristic sound, the probability that its sound is audible. There are 256 units in total for each method.}
  \label{tbl:objsound}
\end{table*}

\section{What does the network learn to detect?}
\label{sec:results}

We evaluate the image representation that our model learned in multiple ways. First, we demonstrate that the internal representation of our model contains convolutional units (i.e., neurons) that are selective to particular objects, and we analyze those objects' distribution.  We then empirically evaluate the quality of the learned representation for several image recognition tasks, finding that it achieves performance comparable to other feature-learning methods that were trained without human annotations.

\label{sec:objdet}Previous work~\citep{zhou2014object} has shown that a CNN trained to predict scene categories will learn convolutional units that are selective for objects -- a result that follows naturally from the fact that scenes are often defined by the objects that compose them.  We ask whether a model trained to predict ambient sound, rather than explicit human labels, would learn object-selective units as well. For these experiments, we used the Clustering variation of our model, because the structure of the network is the same as the scene-recognition model used in \cite{zhou2014object} (whereas the Binary model differs in that it solves a multi-label prediction problem).

\vpar{Labeling object-selective units} Following \citet{zhou2014object}, we visualized the images that each neuron in the top convolutional layer (conv5) responded most strongly to.  To do this, we sampled a pool of 200,000 images from our Flickr video test set.  We then collected, for each convolutional unit, the 60 images in this set that gave the unit the largest activation. Next, we applied the visualization technique of \citet{zhou2014object} to approximately superimpose the unit's receptive field onto the image.  Specifically, we found all of the spatial locations in the layer for which the unit's activation strength was at least half that of its maximum response.  We then masked out the parts of the image that were not covered by the receptive field of one of these high-responding spatial units. We assumed a circular receptive field, obtaining its radius from \citet{zhou2014object}.

We then labeled the neurons by showing the masked images to human annotators on Amazon Mechanical Turk (three per unit), asking them: (1) whether an object is present in many of these regions, and if so, what it is; (2) to mark the images whose activations contain these objects. Unlike \cite{zhou2014object}, we only searched for units that were selective to objects, and did not allow labels for textures or other low-level image structure. For each unit, if at least 60\% of its top 60 activations contained the object in question, we considered it to be {\em selective} for the object (or, following \cite{zhou2014object}, we say that it is a {\em detector} for that object).  We (an author) then assigned an object name to the unit, using the category names provided by the SUN database~\citep{xiao2010sun}.

We found that \numobjunits of the 256 units in our model were object-selective in this way, and we show a selection of them in~\fig{fig:neuronsamples} (additional examples are provided in \fig{fig:allneuronvis}).  In \fig{fig:threshold-curve}, we study how the number of object-selective units changes as we make our evaluation criteria more stringent, by increasing the 60\% threshold.

\xpar{Explaining which objects emerge}
We compared the number of object-selective units to those of a CNN trained to recognize human-labeled scene categories on Places~\citep{zhou2014object}.  As expected, this model -- having been trained with explicit human annotations -- contained significantly more such units (\numplacesunits units).  We also asked whether object-selective neurons appear in the convolutional layers when a CNN is trained on other tasks that do not use human labels. As a simple comparison, we applied the same methodology to the egomotion-based model of \cite{agrawal2015learning} and to the tracking-based method of \cite{wang2015unsupervised}.  We applied these networks to large crops (in all cases resizing the input image to 256$\times$256 pixels and taking the center 227$\times$227 crop), though we note that  they were originally trained on significantly smaller cropped regions.

Do different kinds of self-supervision lead to different kinds of object selectivity?  Using the unit visualization method, we found that the tracking-based model also learned object-selective units, but that the objects that it detected were often textural ``stuff,'' such as grass, ground, and water, and that there were fewer of these detection units in total (\numtrackingunits of 256).  The results were similar for the egomotion-based model, which had \nummotionunits such units.  In \fig{fig:objdistr}, we provide the distribution of the objects that the units were selective to.  We also visualized neurons from the method of \cite{doersch2015unsupervised} (as before, applying the network to whole images, rather than to patches).  We found a significant number of the units were selective for position, rather than to objects.  For example, one convolutional unit responded most highly to the upper-left corner of images -- a unit that may be useful for the training task, which involves predicting the relative position of image patches (moreover, \cite{doersch2015unsupervised} suggests that the model uses low-level cues, such as chromatic aberration, rather than semantics).  In \fig{fig:neuronsamples}, we show visualizations of a selection of object-detecting neurons for all of these methods.

The differences between the objects detected by these methods and our own may have to do with the requirements of the tasks being solved.  The other unsupervised methods, for example, all involve comparing multiple input images or cropped regions in a relatively fine-grained way. This may correspondingly change the representation that the network learns in its last convolutional layer -- requiring its units to encode, say, color and geometric transformations rather than object identities.  Moreover, these networks may represent semantic information in other (more distributed) ways that would not necessarily be revealed through this visualization method.

\makeatletter{}%
\begin{figure}
  \includegraphics[width=\linewidth]{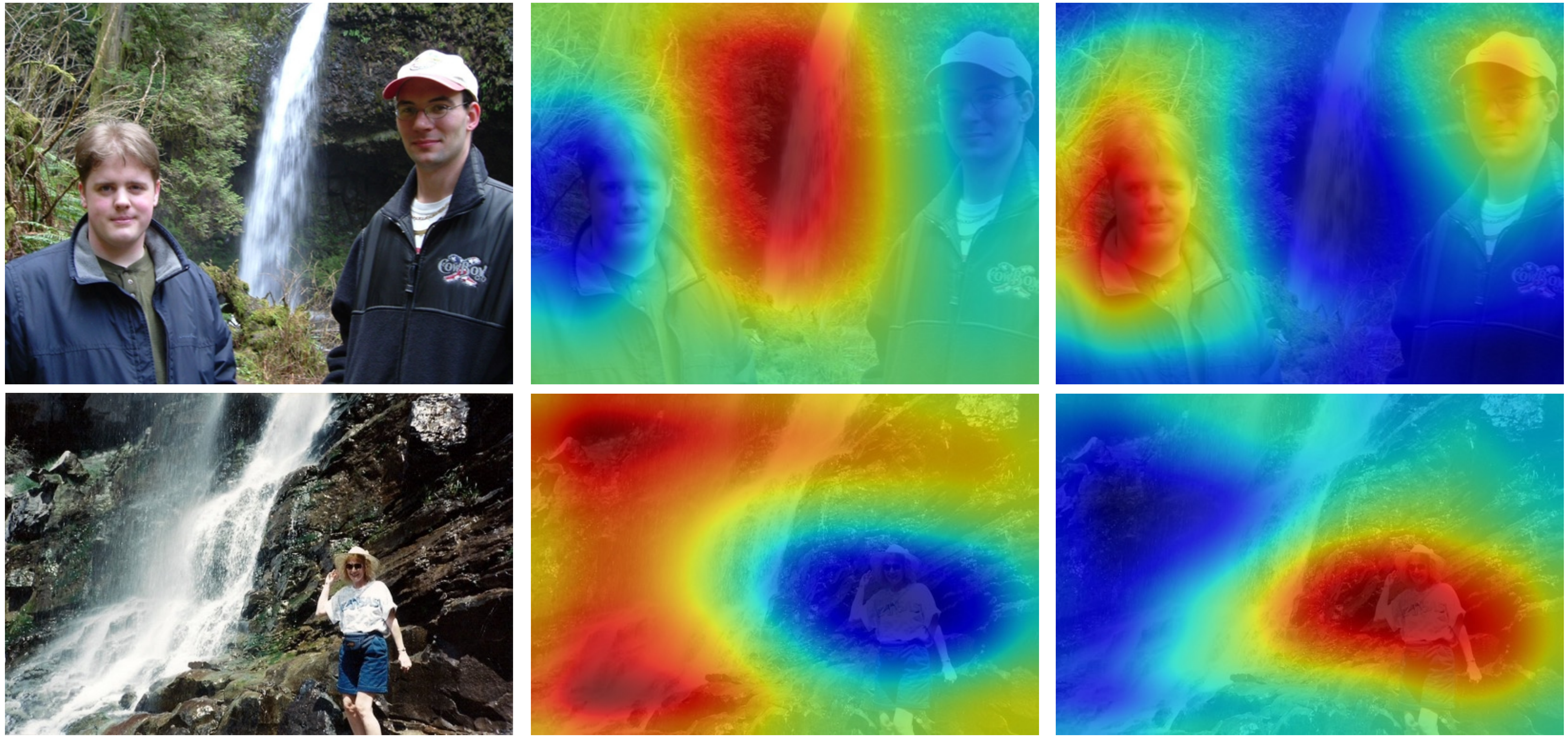}
  \indent \hspace{5mm} Input image \hspace{8mm} Flowing water \hspace{10mm} Speech
  \caption{Class activation maps (CAMs) for speech and flowing water
    sounds. The categories correspond to the third and fifth examples
    in \fig{fig:topcam}. The CAM is colored such that red corresponds
    to high probability of an audio category.}
  \label{fig:multicam}
\end{figure}
\makeatletter{}%
\begin{figure*}[t!]
\centering

\begin{tabular}{c}
Speech\\
\includegraphics[width=1.0\linewidth]{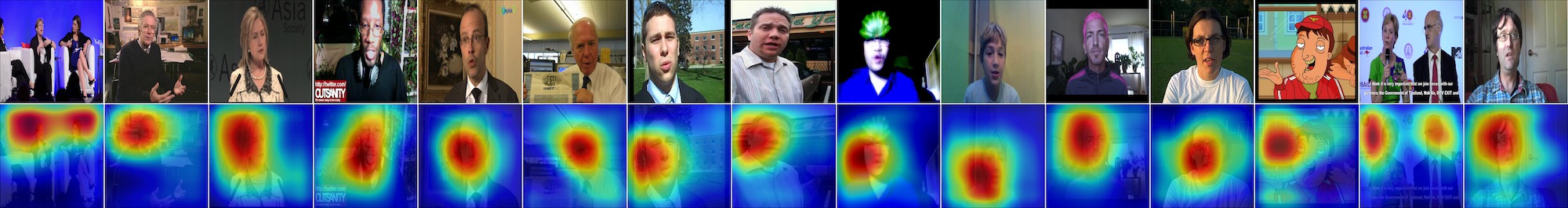}\vspace{1mm}\\
Speech\\
\includegraphics[width=1.0\linewidth]{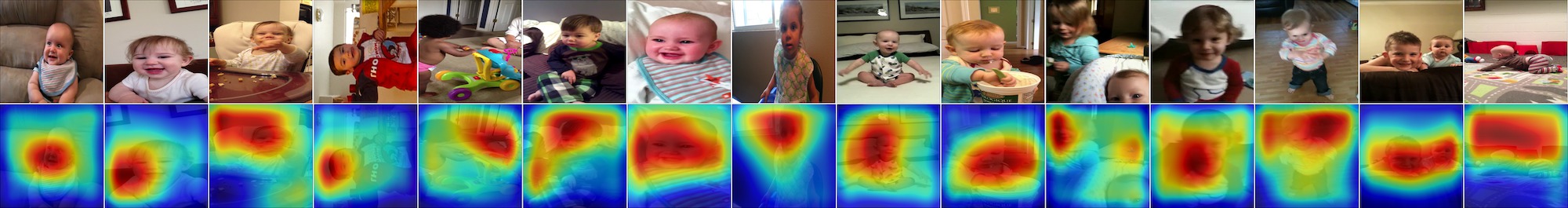}\vspace{1mm}\\
Speech\\
\includegraphics[width=1.0\linewidth]{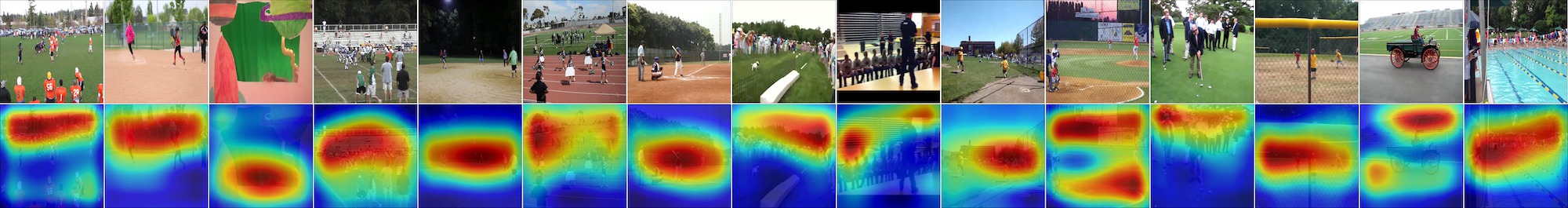}\vspace{1mm}\\
Wind\\
\includegraphics[width=1.0\linewidth]{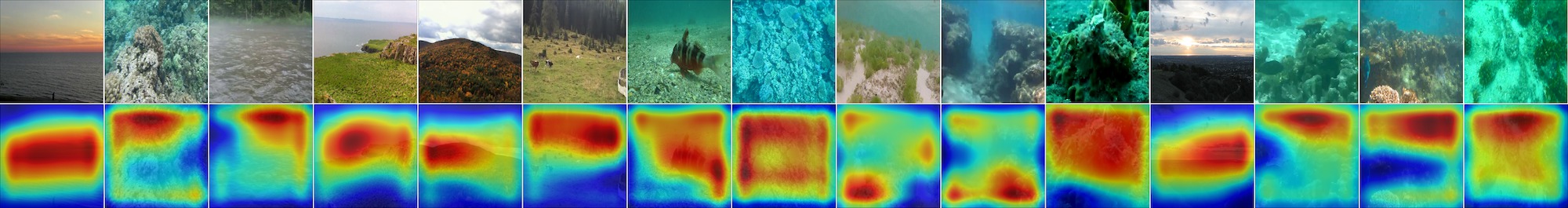}\vspace{1mm}\\
Flowing water \\
\includegraphics[width=1.0\linewidth]{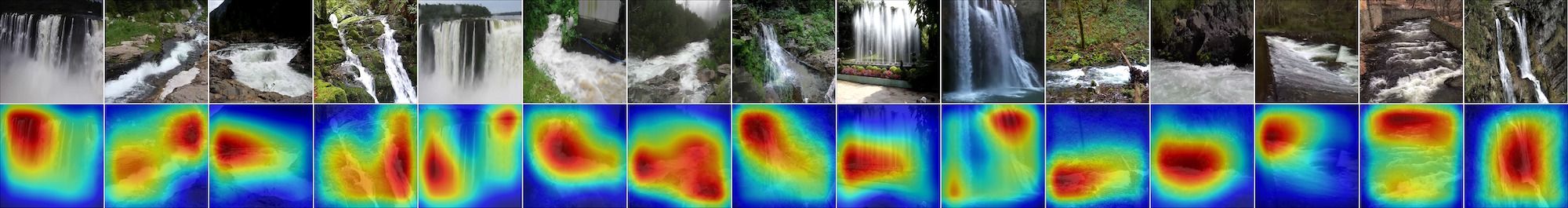}\\
\midrule
 No characteristic sound (mixture of many sounds)\\
 \includegraphics[width=1.0\linewidth]{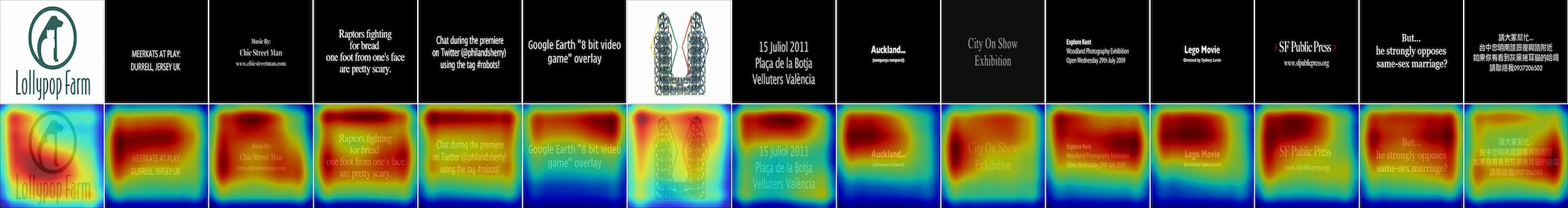}\vspace{1mm}\\
 Silence\\
\includegraphics[width=1.0\linewidth]{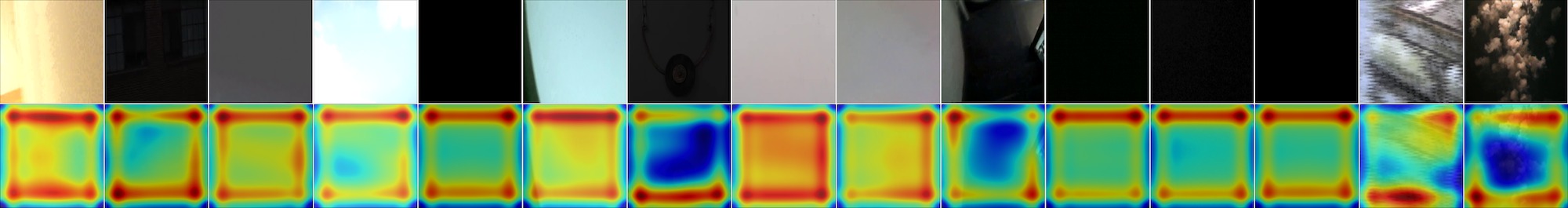}\\
\end{tabular}

\vspace{-10pt}
\caption{For 7 (of 30) audio categories, we show the network's most
  confident predictions and their CAMs, and we provide a description
  of their sound. We show results from the YFCC100m video dataset (to
  avoid having redundant images from similar videos, we show
  for each category at most one example per Flickr user).}
\label{fig:topcam}
\end{figure*}

Next, we asked what kinds of objects our network learned to detect.  We hypothesized that the object-selective neurons were more likely to respond to objects that produce (or are closely associated with) characteristic sounds\footnote{For conciseness, we sometimes call these ``sound-making'' objects, even if they are not literally the source of the sound.}. To evaluate this, we (an author) labeled the SUN object categories according to whether they were closely associated with a characteristic sound in the videos that contained the top detections. We denote these categories with a $*$ in \fig{fig:objdistr}. We found that some objects, such as fireworks, were usually associated in these videos with the sound of wind or human speech, rather the sound of the object itself. We therefore chose not to count these as objects associated with characteristic sounds. Next, we counted the number of units that were selective to these objects, finding that our model contained significantly more such units than a scene-recognition network trained on the Places dataset, both in total number and as a proportion (\tbl{tbl:objsound}). A significant fraction of these units were selective to people (adults, babies, and crowds). 

\xpar{Analyzing the types of objects that were detected} Finally, we asked whether the sounds that these objects make were actually present in the videos that these video frames were sampled from.  To do this, we listened to the sound of the top 30 video clips for each unit, and recorded whether the sound was made by the object that the neuron was selective to (\eg, human speech for the {\em person} category).  We found that \fracvidsound of these videos contained the objects' sounds (\tbl{tbl:objsound}).

To examine the effect of the dataset used to create the neuron
visualizations, we applied the same neuron visualization technique to
200,000 images sampled equally from the SUN and ImageNet datasets (as
in \cite{zhou2014object}). We show examples of these neurons
in \fig{fig:sun-neuron-vis} and plot their distribution \fig{fig:objsun}. As
expected, we found that the distribution of objects was similar to
that of the YFCC100m dataset. However, there were fewer detectors in
total (\numobjsununits vs. \numobjunits), and there were some
categories, such as {\em baby}, that appeared significantly less often
as a fraction of the total detectors.  This may be due to the
differences in the underlying distribution of objects in the
datasets. For example, SUN focuses on scenes and contains more objects
labeled {\em tree}, {\em lamp}, and {\em window} than objects labeled
{\em person} \citep{zhou2014object}. We also computed a detector
histogram for the model of \cite{wang2015unsupervised}, finding that
the total number of detectors was similar to the sound-based model
(\numobjtrackingsununits detectors), but that, as before, the dominant
categories were textural ``stuff'' (e.g., grass, plants).

\subsection{Visualizing sound predictions}
\label{sec:vissoundpred}

These neuron visualizations suggest that our model, internally, is
coding for different object categories, To more directly visualize the
relationship between visual structures and sound categories, we
trained a variation of our model to predict a class activation map
(CAM) \citep{zhou2016cam}.  Following \citet{zhou2016cam}, we replaced
the fully connected layers of our model with convolutions whose
activations are spatially averaged to produce class probabilities (\ie
using global average pooling \citep{lin2013network}).  Under this
model, each spatial position independently casts a vote for each sound
category.  These spatial votes, then, can be used to localize the
visual structures that the network is detecting.

In \fig{fig:topcam} we show, for 7 of 30 audio classes, the images for
which our model assigned the highest probability, along with their
corresponding CAMs (displayed as heat maps).  We also provide a
qualitative description for the audio categories, which we (an author)
obtained by listening to the audio clips that are nearest to its
centroid (similar to \sect{fig:soundclusters}). Together, these
visualizations help to link visual structures with sound. We see, for
example, that speech sounds often correspond to faces, while flowing
water sounds correspond to waterfalls.  In \fig{fig:multicam} we also
show qualitative examples where CAM visualizations from two different
audio categories -- one corresponding to speech, another to flowing
water sounds -- localized two different object types.

\makeatletter{}%
\section{Evaluating the image representation}
\label{sec:feateval}

\makeatletter{}%

\begin{table*}[t!]
  \centering
  \setlength{\tabcolsep}{2.5pt}
  \begin{tabular}{l C{.08\linewidth}C{.08\linewidth}C{.08\linewidth}C{.08\linewidth} C{.08\linewidth}C{.08\linewidth}C{.08\linewidth}C{.08\linewidth} }
        \toprule
        \multirow{2}{*}{Method} & \multicolumn{4}{c}{VOC Cls. (\%mAP)} & \multicolumn{4}{c}{SUN397 (\%acc.)} \\
        \cmidrule(r{2pt}){2-5}  \cmidrule(l{2pt}){6-9}
        & {max5} & {pool5} & {fc6} & {fc7}    & {max5} & {pool5} & {fc6} & {fc7}\\
        \midrule
        Sound (cluster)                                & 36.7 & 45.8 & 44.8 & 44.3                           & {\bf 17.3} & {\bf 22.9} & 20.7 & 14.9   \\
        Sound (binary)                                 & {\bf 39.4} & 46.7 & 47.1 & 47.4                           & 17.1 & 22.5 & {\bf 21.3} & {\bf 21.4}         \\
        Sound (spect.)                                 & 35.8 & 44.0 & 44.4 & 44.4                           & 14.6 & 19.5 & 18.6 & 17.7      \\
        Colorization \citep{zhang2016colorful}         & 38.8    & {\bf 48.3} & {\bf 49.1} & {\bf 51.0}         & 16.0 & 20.3 & 21.2 & 18.4 \\
        Object motion \citep{pathak2016learning}       & 32.4 & 40.8 & 31.5 & 23.6                           & 12.9 & 15.8 & 7.5  & 3.4 \\
        Texton-CNN                                     & 28.9 & 37.5 & 35.3 & 32.5                           & 10.7 & 15.2 & 11.4 & 7.6      \\
        K-means \citep{krahenbuhl2015data}             & 27.5 & 34.8 & 33.9 & 32.1                           & 11.6 & 14.9 & 12.8 & 12.4      \\
        Tracking \citep{wang2015unsupervised}          & 33.5 & 42.2 & 42.4 & 40.2                           & 14.1 & 18.7 & 16.2 & 15.1      \\
        Patch pos. \citep{doersch2015unsupervised}     & 27.7 & 46.7 & - & -                           & 10.0 & 22.4 & - & -     \\
        Egomotion \citep{agrawal2015learning}          & 22.7 & 31.1 & - & -                                 & 9.1  & 11.3 & - & -     \\
        \midrule                                        
        ImageNet \citep{krizhevsky2012imagenet}         & {\bf 63.6} & {\bf 65.6} & {\bf 69.6} & {\bf 73.6}   & 29.8 & 34.0 & 37.8 & 37.8   \\
        Places \citep{zhou2014places}                   & 59.0 & 63.2 & 65.3 & 66.2                           & {\bf 39.4} & {\bf 42.1} & {\bf 46.1} & {\bf 48.8}    \\
        \bottomrule
  \end{tabular}
    \caption{\label{fig:classify} Mean average precision for PASCAL VOC 2007 classification, and accuracy on SUN397. Here we trained a linear SVM using the
    top layers of different networks. We note in \sect{sec:feateval} that the shape of these layers varies between networks.}
\end{table*}

\begin{table*}[t!]
  \centering
  \setlength{\tabcolsep}{6pt}
      \begin{tabular}{lc}
      \toprule
      Method  & (\%mAP) \\
      \midrule
      Random init. \citep{krahenbuhl2015data} & 41.3 \\
      Sound (cluster)  & 44.1 \\ %
      Sound (binary)  & 43.3 \\ %
      Motion \citep{wang2015unsupervised,krahenbuhl2015data} & 47.4 \\
      Egomotion \citep{agrawal2015learning,krahenbuhl2015data} & 41.8 \\
      Patch position \citep{doersch2015unsupervised,krahenbuhl2015data} & 46.6 \\
      Calibration + Patch  \citep{doersch2015unsupervised,krahenbuhl2015data} & {\bf 51.1} \\
      \midrule
      ImageNet \citep{krizhevsky2012imagenet} & {\bf 57.1} \\ %
      Places \citep{zhou2014places} & 52.8 \\ %
      \bottomrule
      \end{tabular}
      \caption{\label{fig:detect} Mean average precision on PASCAL VOC 2007 using Fast-RCNN \citep{girshick2015fast}. We initialized the CNN weights using those of our learned sound models.}
\end{table*}

\begin{table*}[t!]  
    \centering
    \setlength{\tabcolsep}{2.2pt}
    \begin{tabular}{lcccccccccccccccccccc}
\toprule
Method & aer & bk  & brd & bt  & btl & bus & car & cat & chr & cow & din & dog & hrs & mbk & prs & pot & shp & sfa & trn & tv  \\
\midrule

Sound (cluster)                            & 68       & 47       & 38       & 54       & 15       & 45       & 66       & 45       & 42       & 23       & 37       & 28       & 73       & 58       & {\bf 85} & 25       & 26       & 32       & 67       & 42       \\
Sound (binary)                             & 69       & 45       & 38       & 56       & {\bf 16} & {\bf 47} & 65       & 45       & 41       & 25       & 37       & 28       & {\bf 74} & {\bf 61} & {\bf 85} & 26       & 39       & 32       & {\bf 69} & 38       \\
Sound (spect.)                             & 65       & 40       & 35       & 54       & 14       & 42       & 63       & 41       & 39       & 24       & 32       & 25       & 72       & 56       & 81       & {\bf 27} & 33       & 28       & 65       & 40       \\
Colorization                               & {\bf 70} & {\bf 50} & {\bf 45} & 58       & 15       & 45       & {\bf 71} & 50       & 39       & {\bf 30} & 38       & {\bf 41} & 72       & 57       & 81       & 17       & {\bf 42} & {\bf 41} & 66       & 38       \\
Tracking \citep{wang2015unsupervised}      & 67       & 35       & 41       & 54       & 11       & 35       & 62       & 35       & 39       & 21       & 30       & 26       & 70       & 53       & 78       & 22       & 32       & 37       & 61       & 34       \\
Object motion                              & 65       & 39       & 39       & 50       & 13       & 33       & 61       & 36       & 39       & 24       & 35       & 28       & 69       & 49       & 82       & 14       & 19       & 34       & 56       & 31       \\
Patch Pos. \citep{doersch2015unsupervised} & {\bf 70} & 44       & 43       & {\bf 60} & 12       & 44       & 66       & {\bf 52} & {\bf 44} & 24       & {\bf 45} & 31       & 73       & 48       & 78       & 14       & 28       & 39       & 62       & {\bf 43} \\
Egomotion \citep{agrawal2015learning}      & 60       & 24       & 21       & 35       & 10       & 19       & 57       & 24       & 27       & 11       & 22       & 18       & 61       & 40       & 69       & 13       & 12       & 24       & 48       & 28       \\
Texton-CNN                                 & 65       & 35       & 28       & 46       & 11       & 31       & 63       & 30       & 41       & 17       & 28       & 23       & 64       & 51       & 74       & 9        & 19       & 33       & 54       & 30       \\
K-means                                    & 61       & 31       & 27       & 49       & 9        & 27       & 58       & 34       & 36       & 12       & 25       & 21       & 64       & 38       & 70       & 18       & 14       & 25       & 51       & 25       \\

\midrule
ImageNet \citep{krizhevsky2012imagenet} & 79 & {\bf 71} & {\bf 73} & 75 & {\bf 25} & 60 & 80 & {\bf 75} & 51 & {\bf 45} & 60 & {\bf 70} & {\bf 80} & {\bf 72} & {\bf 91} & 42 & {\bf 62} & 56 & 82 & 62\\
Places \citep{zhou2014places} & {\bf 83} & 60 & 56 & {\bf 80} & 23 & {\bf 66} & {\bf 84} & 54 & {\bf 57} & 40 & {\bf 74} & 41 & {\bf 80} & 68 & 90 & {\bf 50} & 45 & {\bf 61} & {\bf 88} & {\bf 63}\\
\midrule
Audio class probability  & 25 & 6 & 12 & 14 & 8 & 6 & 28 & 15 & 21 & 5 & 12 & 15 & 10 & 7 & 75 & 7 & 4 & 9 & 10 & 8\\
\bottomrule
    \end{tabular}
    \caption{\label{fig:classify-class} Per-class AP scores for the VOC 2007 classification task with pool5 features (corresponds to mAP in (a)). We also show the performance obtained using the predicted log-probability of each audio category, using the CAM-based model (\sect{sec:vissoundpred}).}
\end{table*}

We have seen through visualizations that a CNN trained to predict
sound from an image learns units that are selective for objects.  Now
we evaluate how well this representation conveys information about
objects and scenes.

\subsection{Recognizing objects and scenes} 
Since our goal is to measure the amount of semantic information
provided by the learned representation, rather than to seek absolute
performance, we used a simple evaluation scheme.  In most experiments,
we computed image features using our CNN and trained a linear SVM to
predict object or scene category using the activations in the top
layers.

\vpar{Object recognition} First, we used our CNN features for object
recognition on the PASCAL VOC 2007 dataset \citep{everingham2010pascal}.  We trained a one-vs.-rest linear SVM to
detect the presence of each of the 20 object categories in the dataset, using the activations of the upper layers of the network as the feature set (pool5, fc6, and fc7). To help understand whether the convolutional units considered in \sect{sec:objdet} directly convey semantics, we also created a global max-pooling feature (similar to \cite{oquab2015object}), where we applied max pooling over the entire convolutional layer.  This produces a 256-dimensional vector that contains the maximum response of each convolutional unit (which we refer to as {\em max5}). Following common practice, we evaluated the network on a center 227$\times$227 crop of each image (after resizing the image to 256$\times$256), and we evaluated the results using mean average precision (mAP).  We chose the SVM regularization parameter for each method by maximizing mAP on the validation set using grid search (we used $\{0.5^k \mid 4 \leq k < 20\}$).

The other unsupervised (or self-supervised) models in our comparison
\citep{doersch2015unsupervised,agrawal2015learning,wang2015unsupervised,zhang2016colorful,pathak2016learning}
use different network designs.  In particular,
\cite{doersch2015unsupervised} was trained on image patches, so
following their experiments we resized its convolutional layers for
227$\times$227 images and removed the model's fully connected
layers\footnote{As a result, this model has a larger pool5 layer than
  the other methods: 7 $\times$ 7 vs. 6 $\times$ 6. Likewise, the fc6
  layer of \cite{wang2015unsupervised} is smaller (1,024
  dims. vs. 4,096 dims.).}.  Also, since the model of
\cite{agrawal2015learning} did not have a pool5 layer, we added one to
it.  We also considered CNNs that were trained with human annotations:
object recognition on ImageNet \citep{deng2009imagenet} and scene
categories on Places \citep{zhou2014places}.  Finally, we considered
using the $k$-means weight initialization method of
\cite{krahenbuhl2015data} to set the weights of a CNN model (we call
this the {\em K-means} model).

As shown in \tbl{fig:classify}, we found that the overall
best-performing model was the recent colorization method of
\citet{zhang2016colorful}, but that the best-performing variation of
our model (the binary-coding method) obtained comparable performance
to the other unsupervised learning methods, such as
\cite{doersch2015unsupervised}.

Both models based on sound textures (Clustering and Binary)
outperformed the model that predicted only the frequency spectrum.
This suggests that the extra time-averaged statistics from sound
textures are helpful.  In \tbl{fig:classify-class}, we report the
accuracy on a per-category basis for the model trained with pool5
features.  Interestingly, the sound-based models 
outperformed other methods when we globally pooled the conv5 features,
suggesting that the convolutional units contain a significant amount
of semantic information (and are well suited to being used at this
spatial scale).

\vpar{Scene recognition} We also evaluated our model on a scene
recognition task using the SUN dataset \citep{xiao2010sun}, a large
classification benchmark that involves recognizing 397 scene
categories with 7,940 training and test images provided in multiple
splits.  Following \cite{agrawal2015learning}, we averaged our
classification accuracy across 3 splits, with 20 examples per scene
category.  We chose the linear SVM's regularization parameter for each
model using 3-fold cross-validation.  The results are shown in
\tbl{fig:classify}.
    
We found that our features' performance was slightly better than that
of other unsupervised models, including the colorization and
patch-based models, which may be due to the similarity of our learning
task to that of scene recognition. We also found that the difference
between our models was smaller than in the object-recognition case,
with both the Clustering and Binary models obtaining performance
comparable to the patch-based method with pool5 features.

\vpar{Pretraining for object detection} Following recent work
\citep{wang2015unsupervised,doersch2015unsupervised,krahenbuhl2015data},
we used our model to initialize the weights of a CNN-based object
detection system, Fast R-CNN \citep{girshick2015fast}, verifying that
the results improved over random initialization (\tbl{fig:detect}).
We followed the training procedure of \cite{krahenbuhl2015data},
training for 150,000 SGD iterations with an initial learning rate of
0.002. We compared our model with other published results (we report
the numbers provided by \cite{krahenbuhl2015data}). We found that our
model performed significantly better than a randomly initialized
model, as well as the method of~\citet{agrawal2015learning}, but that
other models (particularly~\citet{doersch2015unsupervised}) worked
significantly better.

We note that the network changes substantially during fine-tuning, and
thus the performance is fairly dependent on the parameters used in the
training procedure. Moreover all models, when fine-tuned in this way,
achieve results that are close to those of a well-chosen random
initialization (within 6\% mAP).  Recent
work~\citep{krahenbuhl2015data,mishkin2015all} has addressed these
optimization issues by rescaling the weights of a pretrained network
using a data-driven procedure. The unsupervised method with the best
performance combines this rescaling method with the patch-based
pretraining of~\citet{doersch2015unsupervised}.

\subsection{Audio representation}

Do the predicted audio categories correlate with the
presence of visual objects?  To test this, we used the (log) posterior
class probabilities of the CAM-based sound-prediction network as
(30-dimensional) feature vectors for object recognition. While the
overall performance, unsurprisingly, is low (14.7\% mAP), we found
that the model was relatively good at recognizing people -- perhaps
because so many audio categories correspond to speech.  Its
performance (75\% AP) was similar to that of the much
higher-dimensional pool5 features of the
\citet{doersch2015unsupervised} network (which obtains 78\% AP). We
show the model's per-category recognition accuracy in
\tbl{fig:classify-class}.

\vpar{Sound cluster prediction task} We also asked how well our model learned to
solve its sound prediction task.  We found that on our test set, the clustering-based model (with \numclusters clusters) chose the correct sound label \testclusteracc of the time.  Pure chance in this case is
\clusterpurechance, while the baseline of choosing the most commonly
occurring label is \clustermostcommon.

\begin{figure}[t!]
  \centering
    \hspace{4mm}\includegraphics[width=0.84\linewidth]{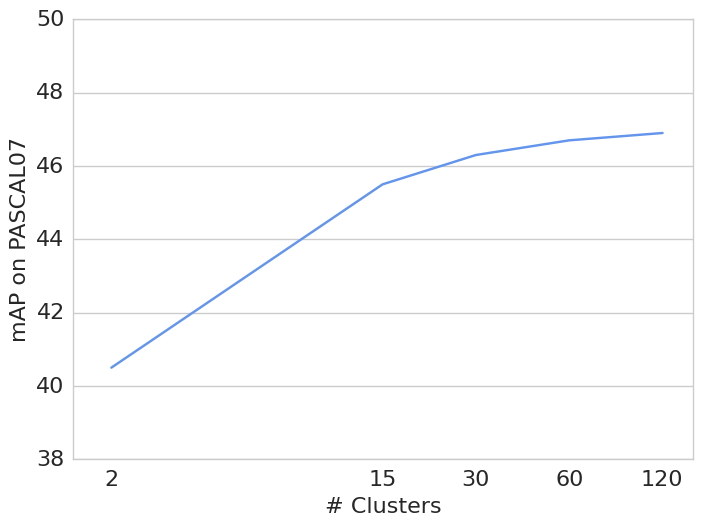} 
    \caption{Object recognition performance (recognition performance
  on PASCAL VOC2007) increases with the number of clusters used to
  define the audio label space.  For our experiments, we used 30
  clusters.} \label{fig:cluster-curve}
\end{figure}

\vpar{Number of clusters} We also investigated how the number of clusters (i.e. audio
categories) used in constructing the audio representation affected the
quality of the visual features.  In \fig{fig:cluster-curve}, we varied
the number of clusters, finding that there is a small improvement from
increasing it beyond \numclusters, and a substantial decrease in
performance when using just two clusters.  We note that, due to the
way that we remove examples whose audio features are not
well-represented by any cluster (\sect{sec:cluster}), the models with
small numbers of clusters were effectively trained with fewer examples
-- a trade-off between cluster purity and data quantity that may
affect performance of these models.

\section{Studying the role of audio supervision}
\label{sec:role}

\makeatletter{}%
\makeatletter{}%

\begin{table}
    \centering
  \begin{tabular}{l c c c c}%
    \toprule 
    \multirow{2}{*}{Method} & \multicolumn{4}{c}{VOC Cls. (\%mAP)}\\
    \cmidrule(r{2pt}){2-5} 
    & {max5} & {pool5} & {fc6 } & {fc7 } \\
    \midrule                             
    Annotations & {\bf 48.8} & {\bf 55.6} & {\bf 56.3} & {\bf 58.1}\\
    Binary & 38.5 & 48.6 & 47.7 & 49.3\\  
    Cluster (30 clusters) & 38.2 & 47.5 & 45.9 & 46.1\\         
    Cluster (120 clusters) & 40.3 & 48.8 & 47.3 & 48.4\\
    \midrule
    ImageNet  & 63.6 & 65.6 & 69.6 & 73.6\\
    \bottomrule
  \end{tabular}                            
  \caption{\label{fig:audioseteval} Comparison between a model trained
    to predict ground-truth sound annotations and our unsupervised
    models, all trained on AudioSet \citep{gemmeke2017audio}.  As in
    \sect{sec:feateval}, we report mean average precision for PASCAL
    VOC 2007 classification, after training a linear SVM on the
    feature activations of different layers.}
\end{table}
\makeatletter{}%
\begin{figure*}[t!]
    \centering
  Training by sound on AudioSet \citep{gemmeke2017audio} (\numaudiosetunits detectors) \\
  \includegraphics[width=\linewidth]{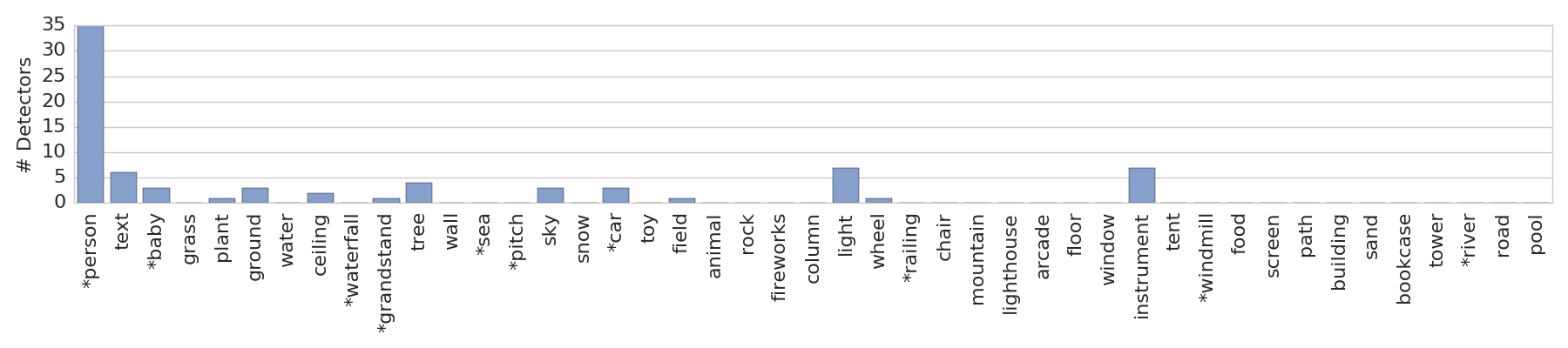}\\
  \caption{We quantify the number of object-selective units for our Cluster method trained on AudioSet~\citep{gemmeke2017audio}.  As before, we visualize the units using the Flickr video dataset (cf. \fig{fig:objdistr}).}
  \label{fig:objdistrsupp}
\end{figure*}

\makeatletter{}%
\begin{figure*}[t!]
\scriptsize\centering
\begin{tabular}{ccc}
instrument&instrument&instrument\\
\includegraphics[width=0.31\linewidth]{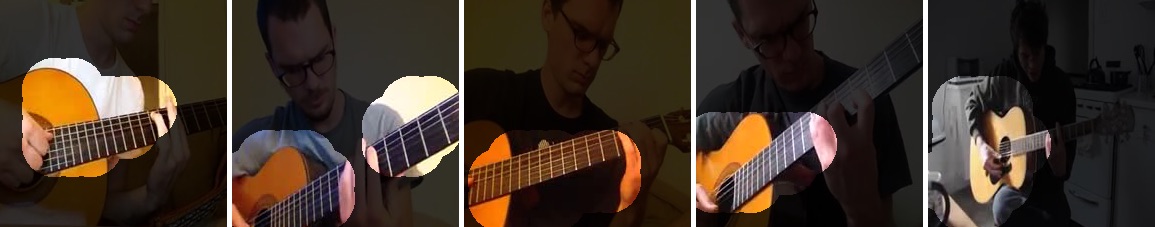}
&\includegraphics[width=0.31\linewidth]{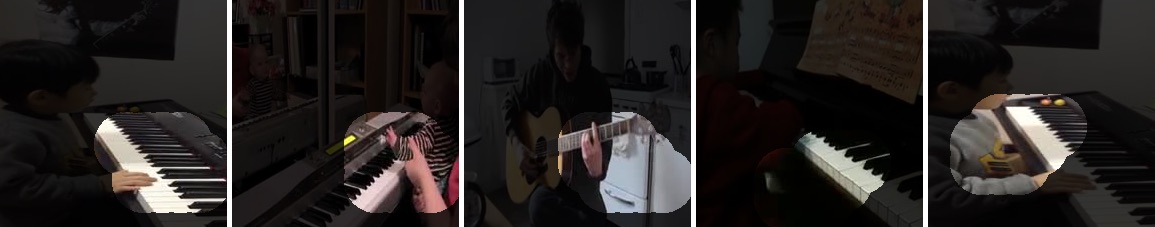}
&\includegraphics[width=0.31\linewidth]{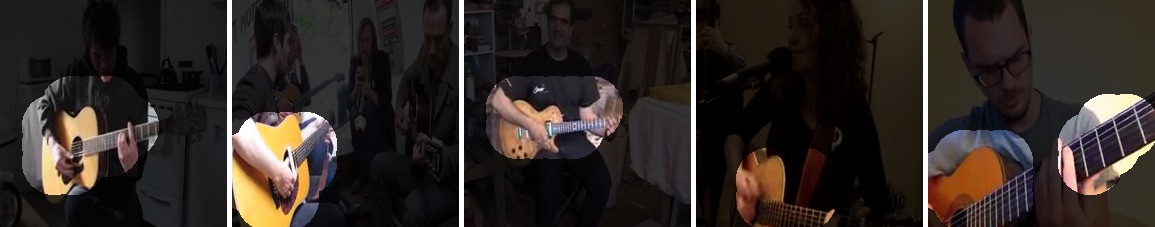}
\end{tabular}
\caption{\label{fig:audiosetunits} Object-selective neurons for a new category (instrument), obtained by training our model on AudioSet~\citep{gemmeke2017audio}. We show the top 5 activations for each unit.}
\end{figure*}

It is natural to ask what role audio plays in the learning process.  Perhaps, for example, our learning algorithm would work equally well if we replaced the hand-crafted sound features with hand-crafted {\em visual} features, computed from the images themselves.  To study this, we replaced our sound texture features with (512-dimensional) visual texton histograms \citep{leung2001representing}, using the parameters from \cite{xiao2010sun}, and we used them to train a variation of our clustering-based model.  

As expected, the images that belong to each cluster are visually coherent, and share common objects.  However, we found that the network performed significantly worse than the audio-based method on the object- and scene-recognition metrics (\tbl{fig:classify}).  Moreover, we found that its convolutional units rarely were selective for objects (generally they responded responded to ``stuff'' such as grass and water).

Likely this large difference in performance is due to the network learning to approximate the texton features, obtaining low labeling error without high-level generalization.  In contrast, the audio-based labels -- despite also being based on another form of hand-crafted feature -- are largely invariant to visual transformations, such as lighting and scale, and therefore predicting them requires some degree of generalization.  This is one benefit of training with multiple, complementary modalities (as explained in \fig{fig:motivation}).

\subsection{Human-annotated sounds} 

Ideally, the audio clustering procedure would produce clusters that map one-to-one with sound sources. In practice, however, the relationship between sound categories and cluster membership is significantly messier (\fig{fig:soundclusters}).

We asked what would happen if, instead of labeling the sound using a clustering procedure, we were to use ``ground-truth'' audio annotations provided by humans.  To study this, we trained a model to predict audio categories from videos in the AudioSet dataset \citep{gemmeke2017audio}.  The videos in this dataset contain 527 sound categories, such as {\em music}, {\em speech}, {\em vehicle}, {\em animal}, and {\em explosion}.  From this dataset, we sampled 962,892 ten-second videos and extracted the middle frame from each one.  We then trained a network (similar in structure to the Binary model) to predict the audio labels. For comparison, we also retrained our audio-based models on this data.

We found, first, that the model that used human annotations performed well (\tbl{fig:audioseteval}): its features performed significantly better than the state-of-the-art unsupervised methods, while still lagging behind ImageNet-based training.  Second, we found that it substantially outperformed our unsupervised models trained on the same dataset. This suggests that there is substantial room for improvement by choosing audio representations that better capture semantics.  

We note that the AudioSet labels may implicitly make use of visual information, both through the use of human annotators (who watched the videos during the labeling process) and due to the fact that visual classifiers were used as an input during the collection process.  As such, the annotations may be best viewed as an upper bound on what is achievable from methods that derive their supervision purely from audio.%

To study the difference between the internal representation of a model learned on (unlabeled) videos in AudioSet, and those of a model trained on the YFCC100m video dataset (Flickr videos), we quantified the object-selective units (\fig{fig:objdistrsupp}). As before, we performed this comparison using the unsupervised {\em Cluster} model.  While there were many similarities between the networks, such as the fact that both have a large number of units tuned to human faces, there are also significant differences.  One such difference is the large number of units that are selective to musical instruments  (\fig{fig:audiosetunits}), which likely emerge due to the large number of music and instrument-related categories in AudioSet.

\makeatletter{}%

\section{Discussion}
\label{sec:discussion}
Sound has many properties that make it useful as a supervisory
training signal: it is abundantly available without human annotations,
and it is known to convey information about objects and scenes.  It is
also complementary to visual information, and may
therefore convey information not easily obtainable from unlabeled
image analysis.

In this work, we proposed using ambient sound to learn visual
representations.  We introduced a model, based on convolutional neural
networks, that predicts a statistical sound summary from a video
frame.  We then showed, with visualizations and experiments on
recognition tasks, that the resulting image representation contains
information about objects and scenes.

Here we considered one audio representation, based on sound textures,
which led to a model capable of detecting certain objects, such as
people and waterfalls. It is natural to ask whether a better audio
representation would lead the model to learn about other objects.
Ideally, one should jointly learn the audio representation with the
visual representation -- an approach taken in recent
work \citep{arandjelovic2017look}. More broadly, we would like to know
which visual objects one can learn to detect through sound-based
training, and we see our work as a step in this direction.
\makeatletter{}%
\vpar{Acknowledgments} This work was supported by NSF grants \#1524817 to A.T;  NSF grants \#1447476 and \#1212849 to W.F.; a McDonnell Scholar Award to J.H.M.; and a Microsoft Ph.D. Fellowship to A.O.  It was also supported by Shell Research, and
by a donation of GPUs from NVIDIA. We thank Phillip Isola for the
helpful discussions, and Carl Vondrick for sharing the data that we
used in our experiments. We also thank the anonymous reviewers for
their comments, which significantly improved the paper (in particular,
for suggesting the comparison with texton features
in \sect{sec:feateval}).

\makeatletter{}%

\appendix
\normalsize
\makeatletter{}%
\begin{figure*}[t!]
\scriptsize\centering
\begin{tabular}{ccc}
person&person&person\\
\includegraphics[width=0.31\linewidth]{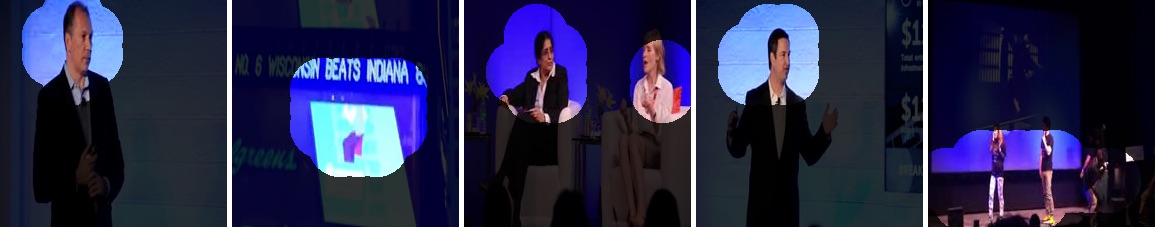}
&\includegraphics[width=0.31\linewidth]{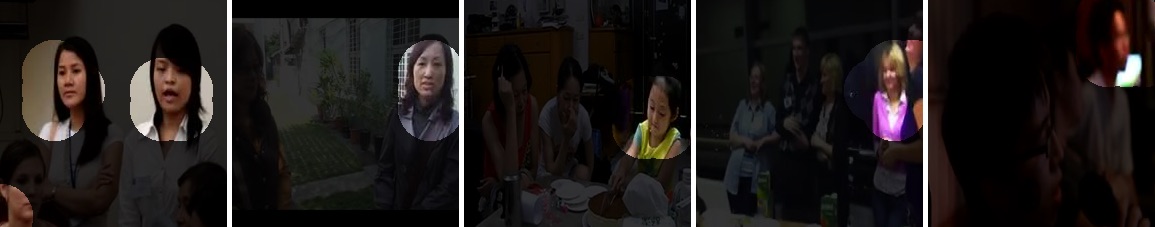}
&\includegraphics[width=0.31\linewidth]{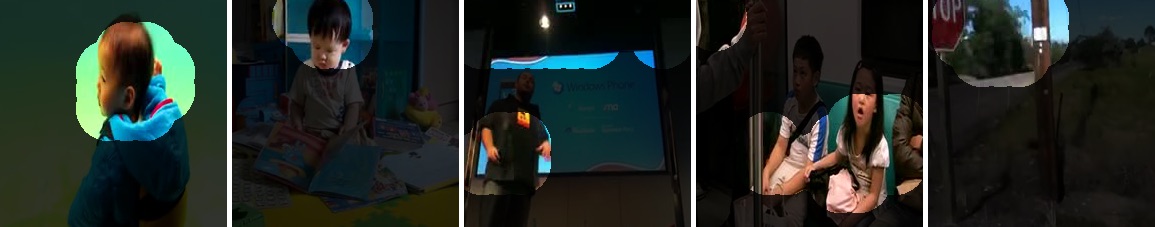}
\\
person&person&person\\
\includegraphics[width=0.31\linewidth]{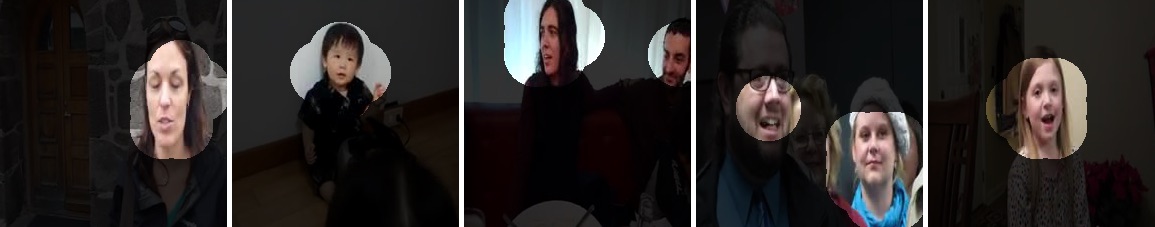}
&\includegraphics[width=0.31\linewidth]{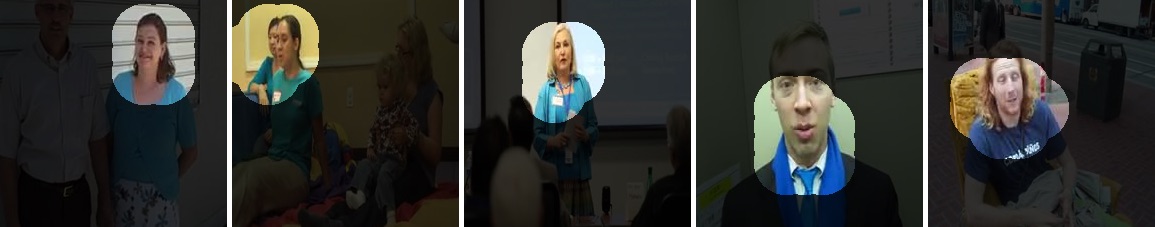}
&\includegraphics[width=0.31\linewidth]{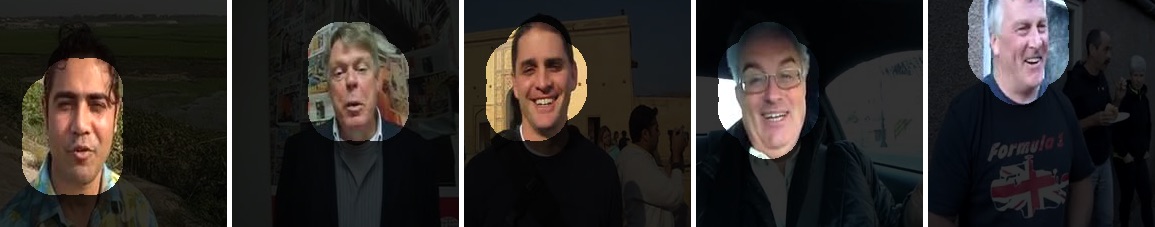}
\\
person&person&person\\
\includegraphics[width=0.31\linewidth]{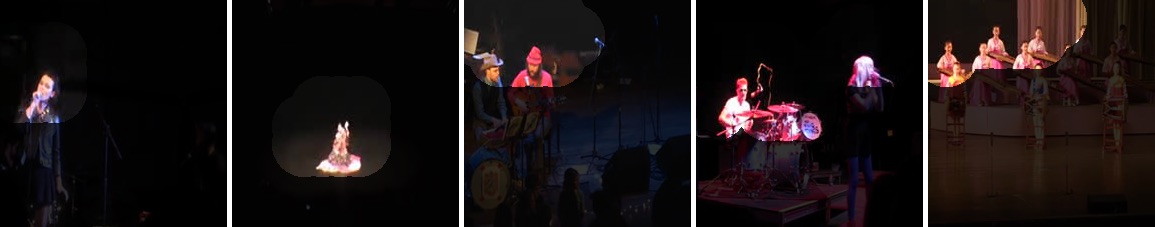}
&\includegraphics[width=0.31\linewidth]{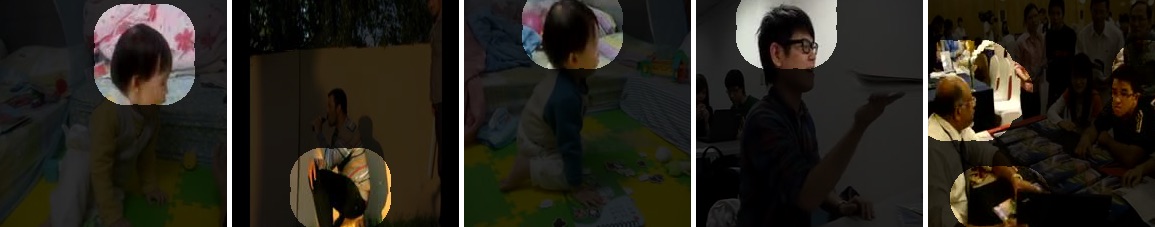}
&\includegraphics[width=0.31\linewidth]{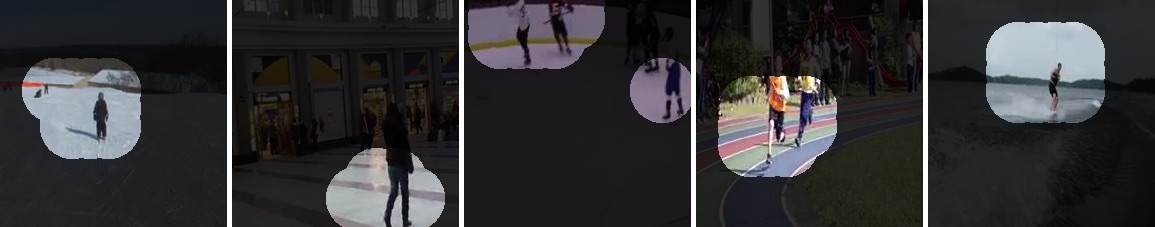}
\\
person&person&person\\
\includegraphics[width=0.31\linewidth]{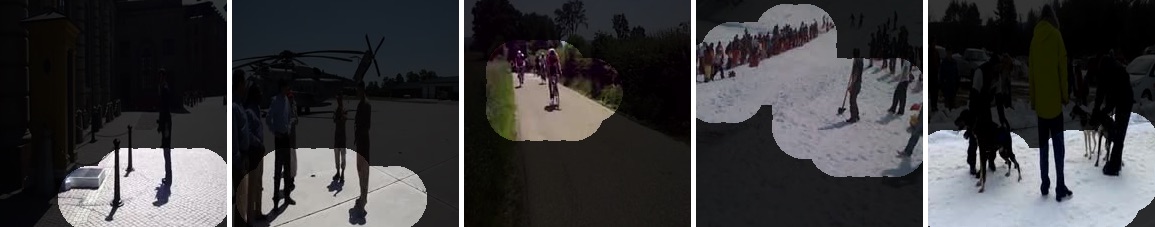}
&\includegraphics[width=0.31\linewidth]{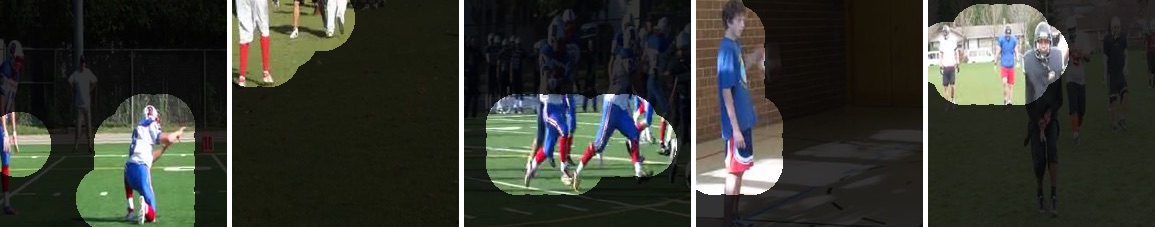}
&\includegraphics[width=0.31\linewidth]{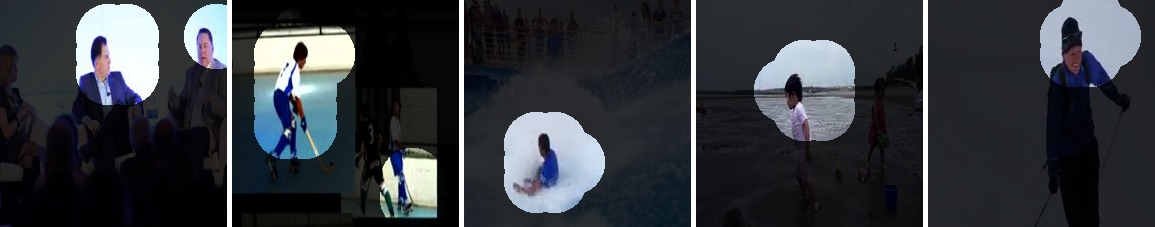}
\\
text&text&text\\
\includegraphics[width=0.31\linewidth]{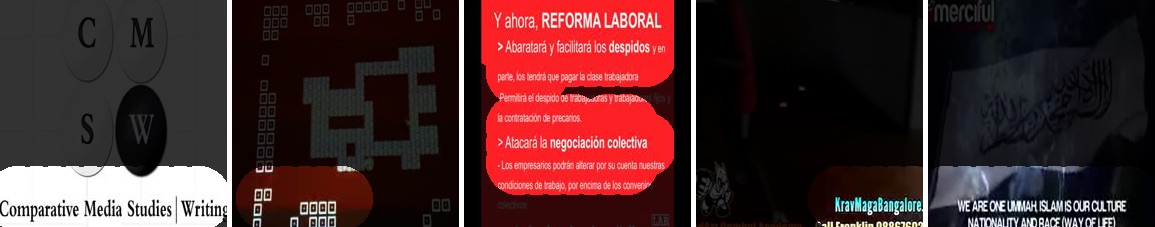}
&\includegraphics[width=0.31\linewidth]{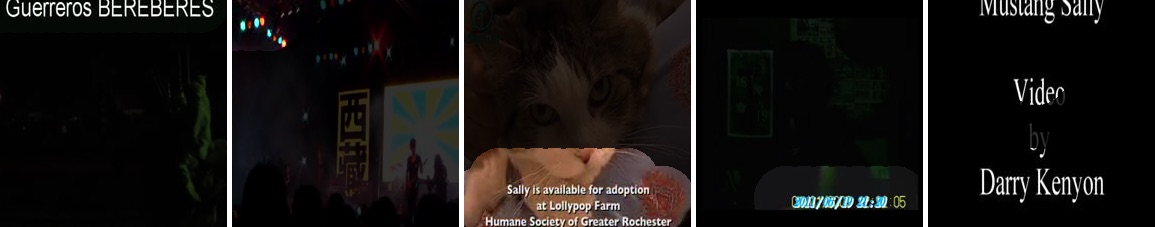}
&\includegraphics[width=0.31\linewidth]{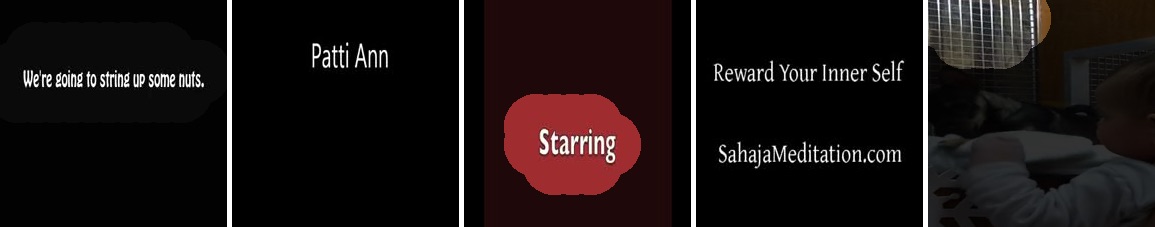}
\\
text&text&text\\
\includegraphics[width=0.31\linewidth]{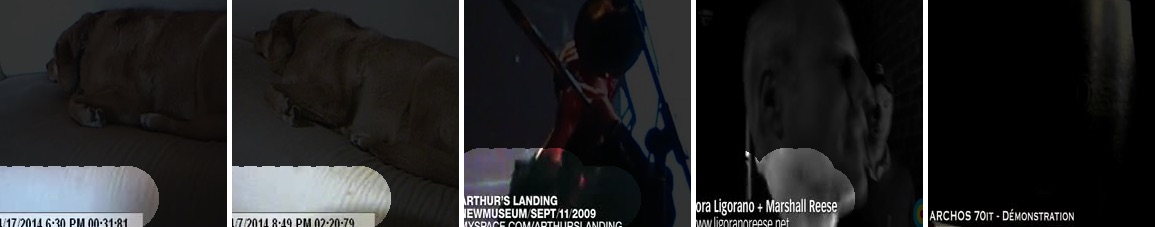}
&\includegraphics[width=0.31\linewidth]{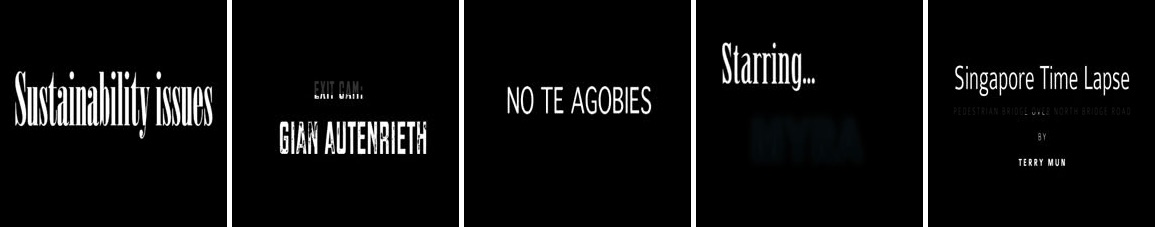}
&\includegraphics[width=0.31\linewidth]{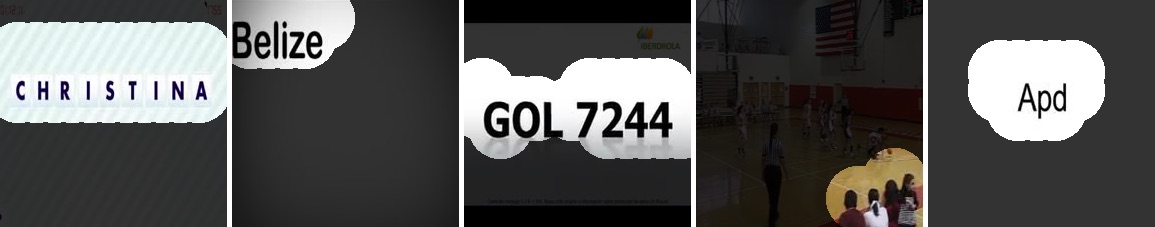}
\\
grass&grass&grass\\
\includegraphics[width=0.31\linewidth]{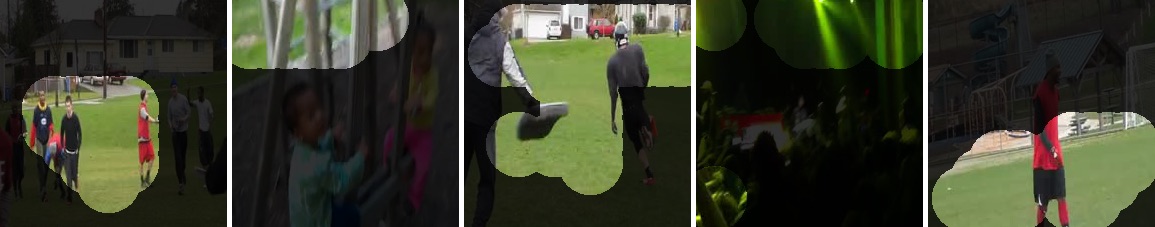}
&\includegraphics[width=0.31\linewidth]{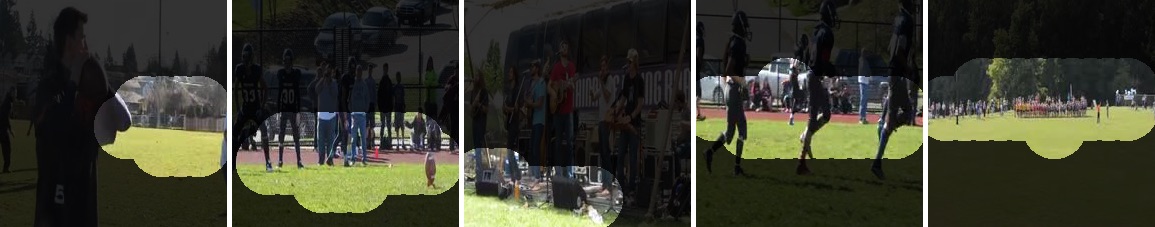}
&\includegraphics[width=0.31\linewidth]{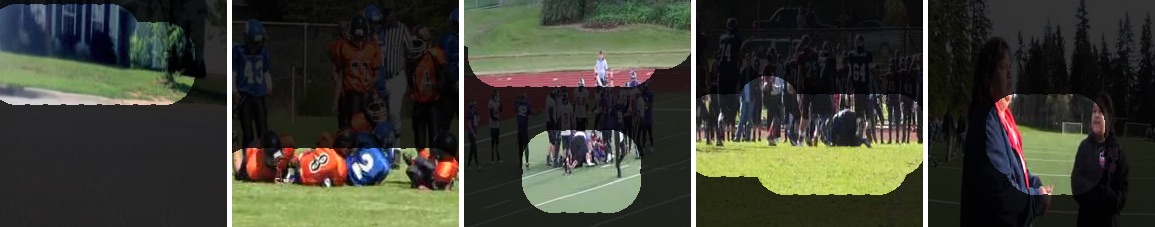}
\\
grass&ceiling&ceiling\\
\includegraphics[width=0.31\linewidth]{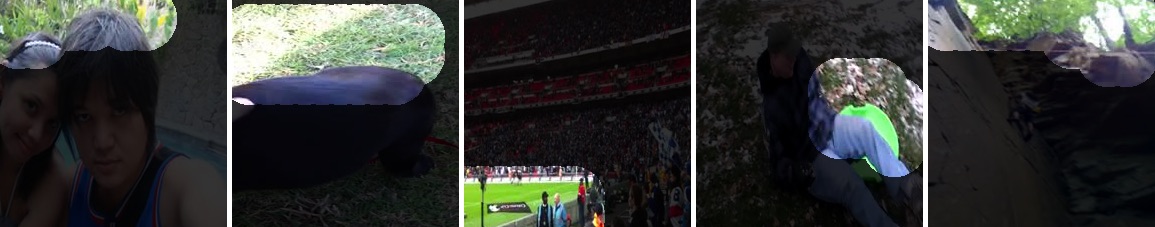}
&\includegraphics[width=0.31\linewidth]{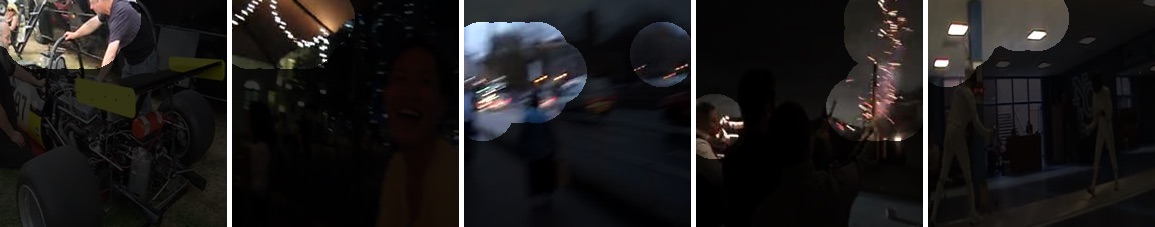}
&\includegraphics[width=0.31\linewidth]{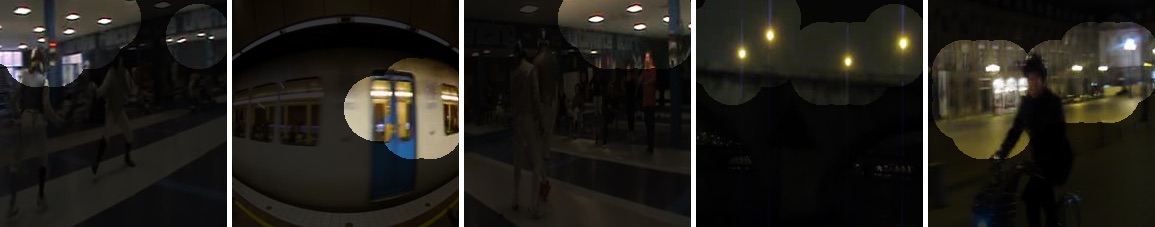}
\\
baby&baby&tree\\
\includegraphics[width=0.31\linewidth]{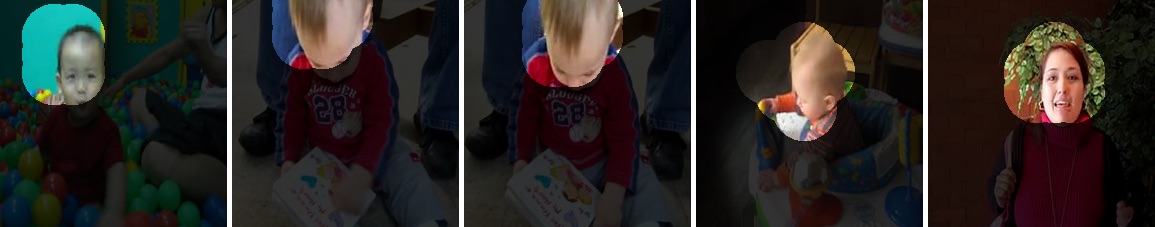}
&\includegraphics[width=0.31\linewidth]{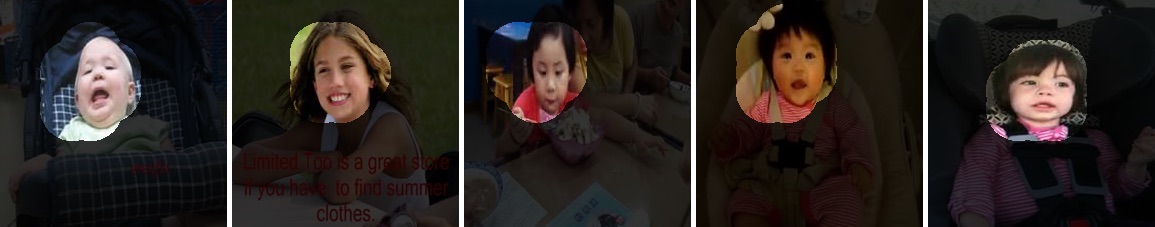}
&\includegraphics[width=0.31\linewidth]{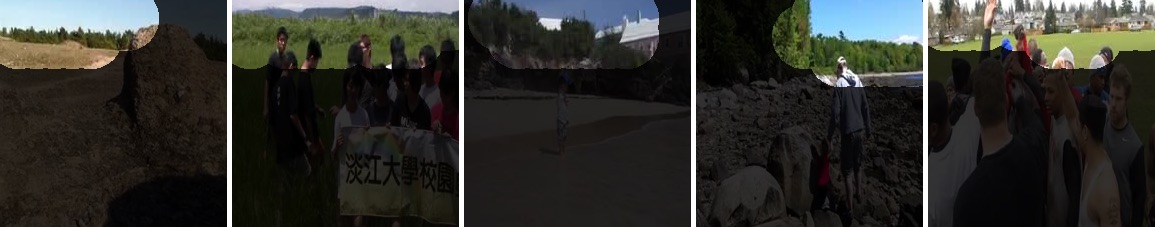}
\\
tree&car&car\\
\includegraphics[width=0.31\linewidth]{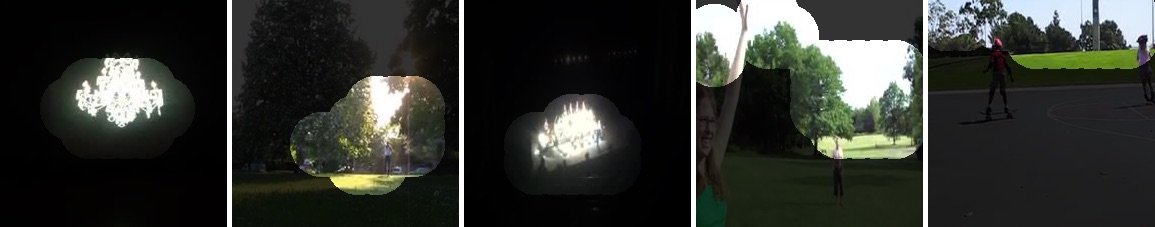}
&\includegraphics[width=0.31\linewidth]{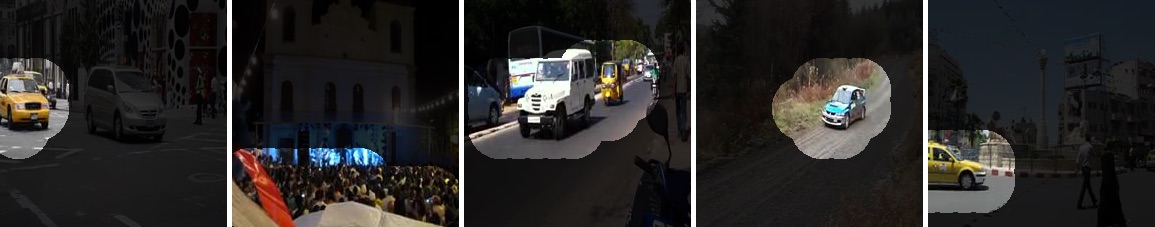}
&\includegraphics[width=0.31\linewidth]{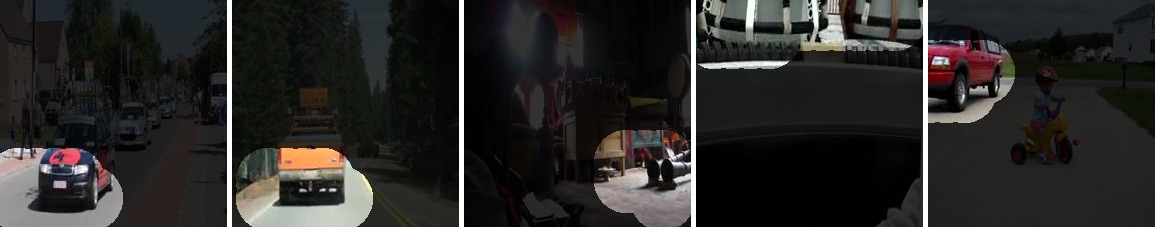}
\\
\midrule
\multicolumn{3}{c}{no object}\\
\includegraphics[width=0.31\linewidth]{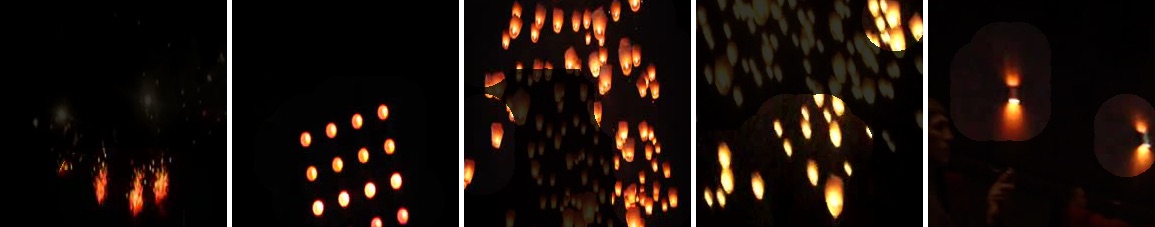}
&\includegraphics[width=0.31\linewidth]{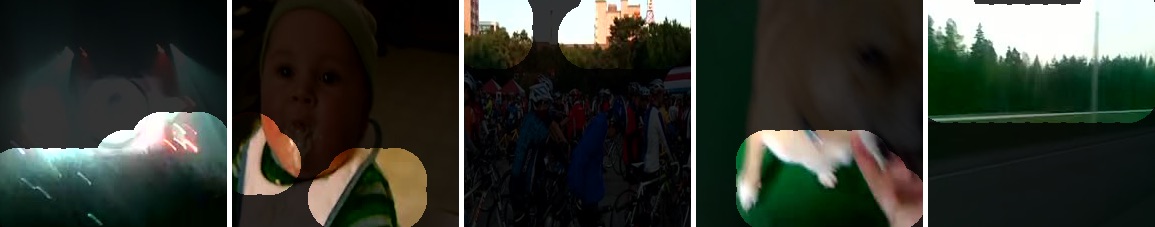}
&\includegraphics[width=0.31\linewidth]{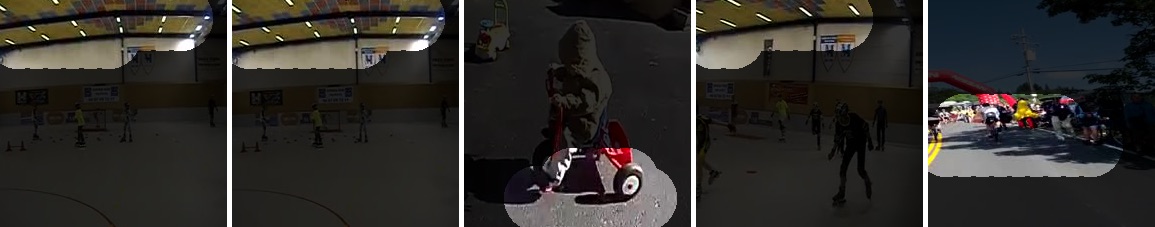}
\\
\multicolumn{3}{c}{no object}\\
\includegraphics[width=0.31\linewidth]{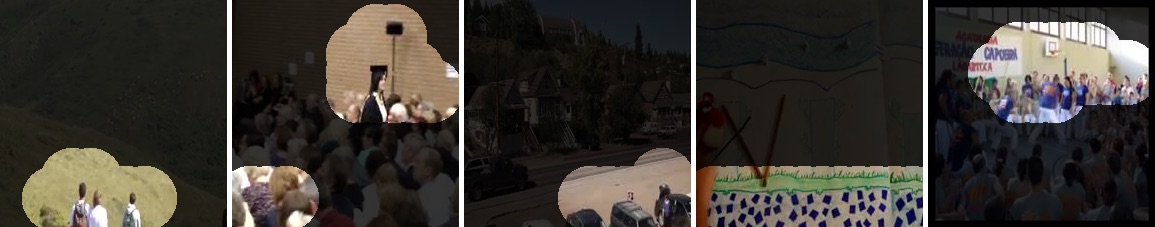}
&\includegraphics[width=0.31\linewidth]{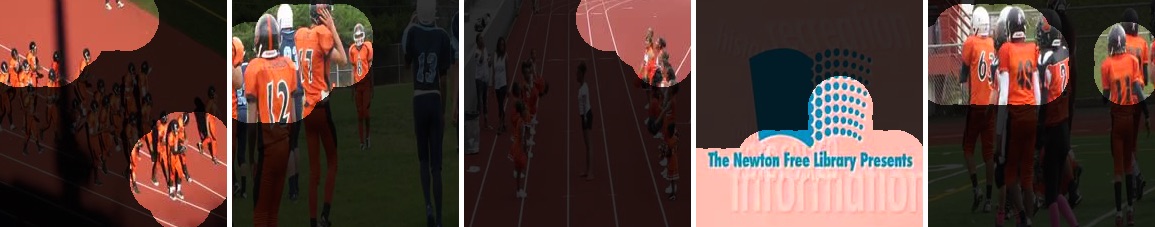}
&\includegraphics[width=0.31\linewidth]{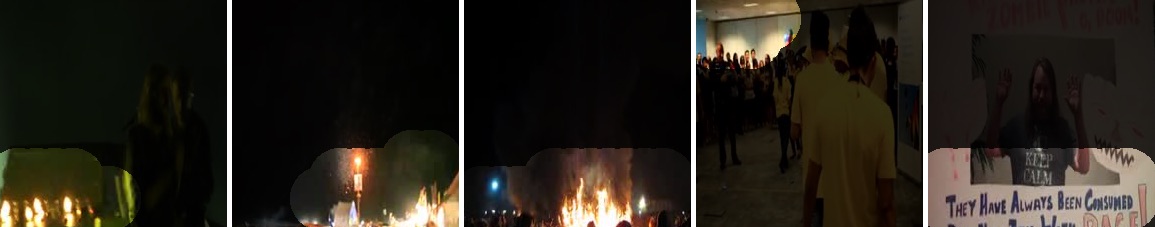}
\\
\end{tabular}
\caption{Additional unit visualizations. We show the top $5$ activations for units in our model ($30$ of \numobjunits from common classes). The last two rows show neurons that were not selective to an object class.}
\label{fig:allneuronvis}
\end{figure*}

\section{Sound textures}
\label{sec:soundtex}
We now describe in more detail how we computed sound textures from audio clips.  For this, we closely follow the work of \cite{mcdermott2011sound}.

\vpar{Subband envelopes} To compute the cochleagram features $\{c_i\}$, we filter the input waveform $s$ with a bank of bandpass filters $\{f_i\}$.  
\begin{equation}
  c_i(t) = |(s \ast f_i) + j H(s \ast f_i)|,
\end{equation}
where $H$ is the Hilbert transform and $\ast$ denotes
cross-correlation. We then resample the signal to 400Hz and compress
it by raising each sample to the $0.3$ power (examples in
\fig{fig:soundtex}).

\vpar{Correlations} As described in Section 3, we compute the correlation between bands using a subset of the entries in the cochlear-channel correlation matrix.  Specifically, we include the correlation between channels $c_j$ and $c_k$ if $|j - k| \in \{1, 2, 3, 5\}$.  The result is a vector $\rho$ of correlation values.

\vpar{Modulation filters} We also include modulation filter responses. To get these, we compute each band's response to a filter bank $\{m_i\}$ of 10 bandpass filters whose center frequencies are spaced logarithmically from 0.5 to 200Hz:
\begin{equation}
  b_{ij} = \frac{1}{N}||c_i \ast m_j||^2,
\end{equation}
where $N$ is the length of the signal. 

\vpar{Marginal statistics}
We estimate marginal moments of the cochleagram features, computing the mean $\mu_i$ and standard deviation $\sigma_i$ of each channel.  We also estimate the loudness, $l$, of the sequence by taking the median of the energy at each timestep, i.e. $l = \mbox{median}(||c(t)||)$.

\vpar{Normalization} To account for global differences in gain, we normalize the cochleagram features by dividing by the loudness, $l$.  Following \cite{mcdermott2011sound}, we  normalize the modulation filter responses by the variance of the cochlear channel, computing $\tilde{b}_{ij} = \sqrt{b_{ij}/\sigma_i^2}$.
Similarly, we normalize the standard deviation of each cochlear channel, computing     $\tilde{\sigma}_{i} = \sqrt{\sigma_{i}^2/\mu_i^2}$.
From these normalized features, we construct a sound texture vector: $[\mu, \tilde{\sigma}, \rho, \tilde{b}, l]$.

\bibliographystyle{spbasic}      %
\bibliography{ambient}   %

\end{document}